\documentclass[lettersize,journal]{IEEEtran}
\usepackage{amsmath,amsfonts}
\usepackage{algorithmic}
\usepackage{algorithm}
\usepackage[table,xcdraw]{xcolor}
\usepackage{array}
\usepackage{makecell} 
\usepackage[caption=false,font=normalsize,labelfont=sf,textfont=sf]{subfig}
\usepackage{textcomp}
\usepackage{booktabs,tabulary}
\usepackage{siunitx}
\usepackage{stfloats}
\usepackage{url}
\usepackage[para,online,flushleft]{threeparttable}
\usepackage{multirow}
\usepackage{mathrsfs} \usepackage{amssymb}
\usepackage{verbatim}
\usepackage{graphicx}
\usepackage{color}
\usepackage{cite}
\usepackage{hyperref}
\usepackage[hyphenbreaks]{breakurl}

\begin{document}

\title{Toward Extremely Lightweight Distracted Driver Recognition With Distillation-Based Neural Architecture Search and Knowledge Transfer}

\author{Dichao Liu, Toshihiko Yamasaki, Yu Wang, Kenji Mase, Jien Kato
\thanks{This manuscript has been accepted for publication as a regular paper in the IEEE Transactions on Intelligent Transportation Systems. Digital Object Identifier 10.1109/TITS.2022.3217342. This work was supported in part by the PhD Professional Toryumon Program, Nagoya University.}

\thanks{\copyright~2022 IEEE. Personal use of this material is permitted. Permission from IEEE must be obtained for all other uses, in any current or future media, including reprinting/republishing this material for advertising or promotional purposes, creating new collective works, for resale or redistribution to servers or lists, or reuse of any copyrighted component of this work in other works.}
\thanks{Dichao Liu was with the Graduate School of Informatics, Nagoya University. He is now with Navier, Inc., Tokyo 102-0084, Japan.~(e-mail:dichao\_liu@outlook.jp).}
\thanks{Toshihiko Yamasaki is with Graduate School of Information Science and Technology, The University of Tokyo, Tokyo 113-8656, Japan.}
\thanks{Yu Wang was with the College of Information Science and Engineering, Ritsumeikan University, Kusatsu-shi, Shiga 525-8577, Japan. He is now with the Center for Information and Communication Technology, Hitotsubashi University, Tokyo 186-8601, Japan.}
\thanks{Jien Kato is with the College of Information Science and Engineering, Ritsumeikan University, Kusatsu-shi, Shiga 525-8577, Japan.}
\thanks{Kenji Mase is with the Graduate School of Informatics, Nagoya University, Nagoya 464-8601, Japan.}
\thanks{}

}




\maketitle

\begin{abstract}
The number of traffic accidents has been continuously increasing in recent years worldwide. Many accidents are caused by distracted drivers, who take their attention away from driving. Motivated by the success of Convolutional Neural Networks (CNNs) in computer vision, many researchers developed CNN-based algorithms to recognize distracted driving from a dashcam and warn the driver against unsafe behaviors. However, current models have too many parameters, which is unfeasible for vehicle-mounted computing. This work proposes a novel knowledge-distillation-based framework to solve this problem. The proposed framework first constructs a high-performance teacher network by progressively strengthening the robustness to illumination changes from shallow to deep layers of a CNN. Then, the teacher network is used to guide the architecture searching process of a student network through knowledge distillation. After that,  we use the teacher network again to transfer knowledge to the student network by knowledge distillation. Experimental results on the Statefarm Distracted Driver Detection Dataset and AUC Distracted Driver Dataset show that the proposed approach is highly effective for recognizing distracted driving behaviors from photos: (\romannumeral1) the teacher network's accuracy surpasses the previous best accuracy; (\romannumeral2) the student network achieves very high accuracy with only 0.42M parameters (around 55\% of the previous most lightweight model). Furthermore, the student network architecture can be extended to a spatial-temporal 3D CNN for recognizing distracted driving from video clips. The 3D student network largely surpasses the previous best accuracy with only 2.03M parameters on the Drive\&Act Dataset. The source code is available at \burl{https://github.com/Dichao-Liu/Lightweight\_Distracted\_Driver\_Recognition\_with\_Distillation-Based\_NAS\_and\_Knowledge\_Transfer}

\end{abstract}


\section{Introduction}
As defined by the National Highway Traffic Safety Administration in the United States (NHTSA), distracted driving is ``any activity that diverts attention from driving''~\cite{national20162015,resalat2015practical}, such as drinking, talking to passengers, etc. Nowadays, distracted driving has become a huge threat to modern society. For example, as reported by the NHTSA, in the United States, traffic accidents caused by distracted driving led to 3,142 or 8.7 percent of all accidents in 2019~\cite{national2019}. 

Recently, Advanced Driver Assistance Systems (ADAS) are being developed to provide technologies that alert the driver to potential problems for preventing accidents. As one of the basic and most important technologies of ADAS, distracted driver recognition (DDR) has attracted much interest from the academic society~\cite{iranmanesh2018adaptive,petermeijer2015vibrotactile,7053946,nemcova2021multimodal}. Many approaches have been developed to use the images taken by a dashcam to recognize whether the driver is driving safely or behaving some categories of distracted driving actions~\cite{abouelnaga2017real,5732698,ai2019double,6518125,baheti2018detection,dhakate2020distracted}. With the effort of the researchers, the recognition accuracy of the DDR task has been increasing, especially when convolutional neural networks (CNNs) are employed in this field~\cite{abouelnaga2017real,ai2019double,zhang_2016}, following the success of CNNs in many other fields. However, the accuracy improvement is generally brought by increased CNN parameter size. The huge parameter size becomes a big problem for real-world applications because of the limitation of vehicle-mounted computing equipment. The purpose of this paper is to design a lightweight and fast network for DDR with high DDR accuracy, which will be very useful for intelligent transportation system (ITS) applications. In the remainder of this section, we start with a review of the existing DDR methods and then briefly present a general overview of our approach.\IEEEpubidadjcol

\subsection{Existing Distracted Driver Recognition Approaches}
Recently, with the success of CNNs in the computer vision field, it has become common to use deep learning models to solve distracted driver recognition (DDR) tasks~\cite{9405644,yan2016driving,abouelnaga2017real}. For example, Yan \textit{et al}.~\cite{yan2016driving} embedded local neighborhood operations and trainable feature selectors within a deep CNN, and by doing so, meaningful features could be selected automatically to recognize distracted drivers.

However, the introduction of CNNs causes the problem of huge parameter size. There are some recent lightweight networks designed for general-purpose computer vision, such as MobileNet~\cite{howard2017mobilenets}, MobileNetV2~\cite{8578572} and SqueezeNet~\cite{iandola2016squeezenet}. However, these lightweight networks are not specifically designed for DDR, and therefore is still room for improvement regarding DDR accuracy and the number of parameters.

There are now also some lightweight networks designed specifically for DDR by hand. For example, Baheti \textit{et al}.~\cite{baheti2020towards} propose the MobileVGG, which reduces the number of parameters by replacing the traditional convolution in the classical VGG structure with depth-wise convolution and point-wise convolution. D-HCNN~\cite{qin2021distracted} is another example, which uses an architecture containing four convolution blocks with the filters of rather large spatial sizes and achieves high performance with small number of filters. However, these networks were designed entirely by hand based on experience with networks used for general-purpose computer vision tasks, so the potential of the network structure could not be reached to the maximum extent possible. Moreover, D-HCNN requires histogram of oriented gradients (HOG)~\cite{dalal2005histograms} in addition to RGB image as the input. HOG counts occurrences of gradient orientation in localized portions of an image and describes the appearance and shape of the local objects. The computation of HOG requires extra processing effort and is not favorable for real-world applications.

In this work, we search for an optimal architecture for the DDR task, which has less parameter size and higher accuracy than the above studies. Our approach is designed by NAS rather than totally by hand and only requires RGB images as inputs.

\subsection{A Brief Overview of the Proposed Approach}

To solve this problem, we propose a distillation-based neural architecture search and knowledge transfer framework. Overall, the proposed framework is based on knowledge distillation~\cite{gou2021knowledge}, which refers to the process of transferring knowledge from a large model (teacher network) to a smaller one (student network). The proposed framework includes three steps: (\romannumeral1) constructing a strong teacher network; (\romannumeral2) searching and define the architecture of a student network under the supervision of the teacher network; (\romannumeral3) transferring the knowledge from the teacher network to the student network.

\begin{figure}[t!]
\centering\includegraphics[width=\linewidth]{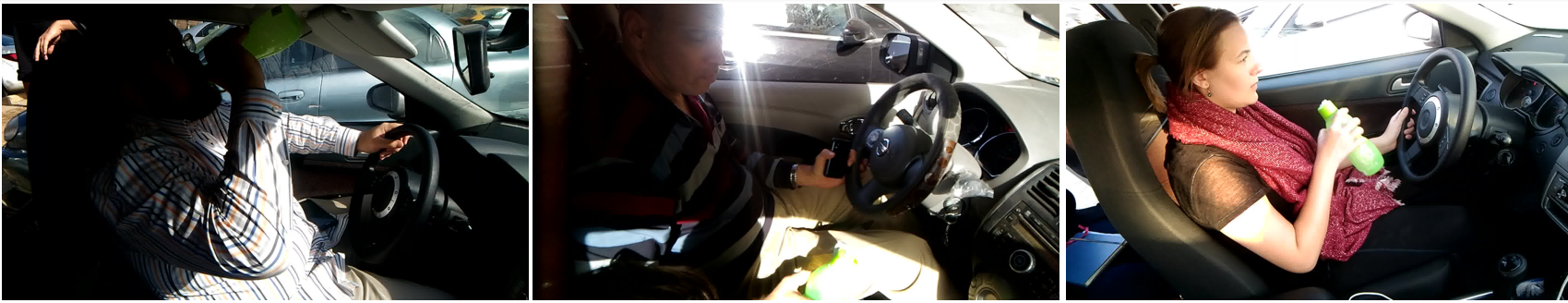}
\caption{Examples of images taken by a camera monitoring the driver's behavior under different illumination conditions. The ground-truth label of the images is ``Drink''. The images are from the AUC Distracted Driver Dataset~\cite{alotaibi2019distracted}.}
\label{fig:illumination}
\end{figure}

\textbf{Teacher Network.} The teacher network is built based on progressive learning (PL). PL is a training strategy that starts the training from shallow layers and then progressively deepens the model by adding new layers to the model~\cite{9405644,yan2016driving,abouelnaga2017real}. In some studies, PL is also regarded as partitioning a network into several segments and progressively training the segments from shallow to deep~\cite{du2020fine,zhao2021pmg}. Progressive learning (PL) was originally proposed for generative adversarial networks~\cite{ahn2018image}. It started with low-resolution images, and then progressively increased the resolution by adding layers to the networks. For example, Wang \textit{et al}.~\cite{wang2018fully} proposed to progressively cascade residual blocks to increase the stability of processing extremely low-resolution images with very deep CNNs. Shaham \textit{et al}.~\cite{shaham2019singan} proposed to reconstruct high-resolution images by a progressive multi-scale approach that progressively up-sample the output from the previous level. Recently, PL has been also applied in fine-grained image classification tasks. For example, Du \textit{et al}.~\cite{du2020fine} and Zhao \textit{et al}.~\cite{zhao2021pmg} used PL to fuse information from previous levels of granularity and aggregate the complementary information across different granularities. 

In this work, we introduce PL into DDR to solve the problem caused by various illumination conditions, such as sunlight and shadow. As shown in Figure~\ref{fig:illumination}, in the real world, the dashcam commonly records the driver's behavior in different illuminations, while the color itself is susceptible to the influence of illumination. RGB information changes considerably under different illuminations, which causes strong intra-class variance in the DDR task. Such intra-class variance affects CNNs from shallow to deep layers. The shallow layers of CNN tend to learn basic patterns, such as different orientations, parallel lines, curves, circles, etc., while the deep layers tend to encode the patterns learned by shallow layers to capture more semantically meaningful information, such as hands, body, etc~\cite{zeiler2014visualizing}. Following the experience learned with bright illumination on what basic patterns are discriminative, the shallow layers of a CNN might fail to find enough discriminative basic patterns in the shadows. 

In this work, we progressively train the teacher network for several stages. During the stages, the training starts from shallow layers and progressively goes deeper with random brightness augmentation~\cite{imgaug} to increase the robustness to the illumination of the layers from shallow to deep. Thereafter, we use the original image to train the aggregation of the models of all stages, considering that the random brightness augmentation might lose some visual information.

\textbf{Student Network.} 
The student network is a compact network that should be able to achieve high recognition performance. This leads to a research question: how to define the architecture of the student network to make it compact, lightweight, yet powerful for DDR, by utilizing the knowledge of the teacher network as supervision? 

To answer this question, we turn our eyes to neural architecture search (NAS). NAS refers to the process of automating architecture engineering to learn a network topology that can achieve best performance on a certain task~\cite{elsken2019neural,zoph2018learning,bashivan2019teacher}. 

The major components of NAS includs searching space, searching algorithm, and evaluation strategy~\cite{elsken2019neural}. With the prior knowledge about typical properties of architectures, NAS approaches commonly define the searching space as a large set of operations (e.g., convolution, fully-connected, and pooling). Each possible architecture in the searching space is evaluated by a certain evaluation strategy~\cite{elsken2019neural,zoph2018learning} and the searching process is controlled by certain searching algorithms, such as reinforcement learning~\cite{zoph2018learning,liu2018progressive,pham2018efficient}, evolutionary search~\cite{real2019regularized}, differentiable search~\cite{DARTS}, or other learning algorithms~\cite{Kirthevasan2018,bashivan2019teacher,wang2021teacher,NEURIPS2018_933670f1}. NAS commonly defines a searching space at first and then uses a certain policy to generate a sequence of actions in the searching space to specify the architecture. 

In this work, we propose a new searching approach for DDR based on the characteristics of the images in the DDR task. We introduce how we define the searching space and the searching strategy as described below.

\textbf{Searching Space.} The images in the DDR task have less diversity and much stronger inter-class similarity than those in many other image recognition tasks. For example, in the fine-grained image recognition task of CUB Birds~\cite{WelinderEtal2010}, the images contain the birds of different species, the background of different habitats, etc. However, in the DDR task, almost all the images can be roughly described as ``a human is driving.'' Thousands of images showing different driving behaviors might be performed by the same person, and the backgrounds of all the images are actually the interior of the same car. 

Due to the above reason,  a large proportion of the visual information does not provide discriminative clues in the DDR task. For example, in CUB Birds, the color of wings, the shape of heads, etc. all provide useful information. Sometimes, even the background provides useful information as a bird image with the sea as the background highly likely shows a certain sea bird. In contrast, in the DDR task, the color of the driver's clothes, the shape of the driver's glasses or hat, almost all the background, etc. are useless information. 

Consequently, the models for the DDR task do not need a huge number of object detectors. The key is to explore some discriminative objects, which are quite universal among different driving behaviors, such as hands, body pose, steering wheel, etc. In CNNs, depth influences the flexibility, and each channel of the filters acts as an object detector~\cite{zeiler2014visualizing}. Thus, the architecture for DDR does not require a very deep structure and a huge number of channels. The above claim is backed up by some earlier observations that the architecture of a decreased number of layers and channels can achieve good results in DDR~\cite{baheti2020towards,qin2021distracted}.

On the other hand, the architecture for DDR must be able to effectively find and capture useful clues from the limited discriminative objects, which is very difficult because: (\romannumeral1) the inter-similarity is strong; (\romannumeral2) the key objects vary largely in size (e.g., hands and body). In this work, we introduce pyramidal convolution (PyConv)~\cite{duta2020pyramidal} into the DDR task. In a standard convolution layer, all the filters have the same spatial size. In contrast, a PyConv layer uses convolution filters of different spatial sizes, and the filters are possible to divide into several groups. Thus, PyConv has very flexible receptive fields, which is beneficial to capture key objects of different sizes. Also, due to its flexibility, PyConv provides a large pool of potential network architectures. In this work, the main searching space is defined as the candidate combinations of filters' spatial sizes and the number of groups. Moreover, the pooling method applied in the model also influences the performance of capturing key objects~\cite{yu2014mixed}. We also search whether to use max pooling or average pooling in the layers.

\textbf{Searching Strategy.} Most of the NAS methods train the possible candidate networks one by one, and evaluate the performance of the trained candidate networks on a validation set~\cite{elsken2019neural,9352235}. The evaluation results are used as metrics to update the architecture searching process. However, the process of candidate evaluation could be very expensive in terms of time, memory, computation, etc. In this work, since we have already constructed a powerful teacher network, we directly use the teacher network to guide the searching. Specifically, we first build a super student network that aggregates all the candidates with a weighted sum, whose weights are regarded as the possibility of choosing each candidate. Then the super student network is trained to learn from the teacher network by knowledge distillation. After the training, the candidates with the maximum weight are chosen to build the architecture of the student network.

After defining the architecture of the student network, the teacher network is utilized again to transfer knowledge to the student network.

Our contributions are summarized as follows:

\begin{itemize}
\item[-] We propose a novel framework for solving the DDR task with high accuracy and a small number of parameters. The research question is solved by the proposed searching strategy. 

\item[-] We mainly carried out the experiments of training the teacher network, defining the student network, and evaluating the performance of the teacher and student networks on two image-based DDR datasets, namely the AUC Distracted Driver Dataset (AUCD2)~\cite{alotaibi2019distracted} and Statefarm Distracted Driver Detection Dataset (SFD3)~\cite{statefarm}. The experimental results show that the teacher network achieves 96.35\% on the AUCD2 and 99.86\%--99.91\% in different splitting settings on the SFD3 with 44.62M parameters, which outperforms the previous state-of-the-art approaches on both datasets. Note that the previous best approach on AUCD2 requires 140M parameters.

\item[-] The student network achieves 95.64\% on the AUCD2 and 99.86\%--99.91\% in different splitting settings on the SFD3  with only 0.42M parameters.

\item[-] The student network architecture can be extended into a spatial-temporal 3D convolutional neural network by replacing the 2D layers with spatial-temporal 3D layers~\cite{hara2018can,tran2015learning,qiu2017learning,carreira2017quo}. We carried out comprehensive experiments in all the tasks of the Drive\&Act Dataset (DAD)~\cite{drive_and_act_2019_iccv}, which is a video-based DDR dataset. The 3D student network is 0.89\%--29.00\% higher than the previous best accuracy in the validation set and 2.05\%--30.88\% higher than the previous best accuracy in the test set. The 3D student network requires only 2.03M parameters.

\end{itemize}

\section{Details of the Proposed Approach}
\begin{algorithm}[t!]
\caption{Building the teacher network based on progressive learning}
\label{alg:algorithm1}
\textbf{Require}: Given a dataset $\mathcal{D}=\{({\rm input}^i,~p_{\rm truth}^i)\}_{i=1}^{I}$ ($I$ is the total number of images in $\mathcal{D}$), and $N$ stages $\{s_1, s_2, ..., s_n, ..., s_N\}$ of the backbone feature extractor $E$.
\begin{algorithmic}[1] 
\FOR{epoch $\in$ [1, num\_of\_epoch]}
\FOR{($\rm input$, $p_{\rm truth}$) in $\mathcal{D}$}
\FOR{$n$ $\in$ [1, $N$]}
\STATE ${\rm input_n} = {\rm Brightness\_augmentor}({\rm input})$
\STATE $x_n$ = $s_n({\rm input}_n)$
\STATE $v_n$ = $\phi_n(x_n)$
\STATE $p_n$ = $\psi_n(v_n)$
\STATE $\mathcal{L}_n$ = $\mathcal{L}_{\rm cls}(p_n,p_{\rm truth})$ 
\STATE \textbf{BACKPROP}($\mathcal{L}_n$)
\ENDFOR
\FOR{$n$ $\in$ [1, $N$]}
\STATE ${\rm input}_n = {\rm input}$
\STATE $x_n$ = $s_n({\rm input}_n)$
\STATE $v_n$ = $\phi_n(x_n)$
\ENDFOR

\STATE $v_{N+1}$ = $f^{\rm concat}$($v_1$, $v_2$, ..., $v_N$)
\STATE $p_{N+1}$ = $\psi_{N+1}(v_{N+1})$
\STATE $\mathcal{L}_{N+1}$ = $\mathcal{L}_{\rm cls}(p_{N+1},p_{\rm truth})$ 
\STATE $\textbf{BACKPROP}$($\mathcal{L}_{N+1}$)
\ENDFOR
\ENDFOR
\end{algorithmic}
\end{algorithm}

\subsection{Teacher Network Construction}
In this subsection, we introduce the details of the teacher network. Let $E$ be the backbone feature extractor, which can be based on any state-of-the-art models, such as SKResNeXt50~\cite{li2019selective}, etc. The layers of $E$ are divided into $N$ segments $\{m_1, m_2, ..., m_n, ..., m_N\}$. Assume $\{s_1, s_2, ..., s_n, ..., s_N\}$ be $N$ consecutive stages from shallow to deep. At each stage of $\{s_1, s_2, ..., s_n, ..., s_N\}$, the training always starts from the first layer of $E$. From $s_1$ to $s_N$, the training gradually goes deeper and covers more layers of $E$. That is, the segments under training at stage $s_n$ are $m_1+m_2+...+m_n$. Let $\{x_1, x_2, ..., x_n, ..., x_N\}$ denote the the output feature maps at $\{s_1, s_2, ..., s_n, ..., s_N\}$. Let $x_n\in\mathbb{R}^{H_n\times W_n \times C_n}$ denote the output feature map at the stage $s_n$, and $H_n$, $W_n$, and $C_n$ respectively denotes the height, width, and the number of channels of $x_n$. We use a set of operations $\{\phi_1(.), \phi_2(.), ..., \phi_n(.), ..., \phi_N(.)\}$ to respectively process $\{x_1, x_2, ..., x_n, ..., x_N\}$ into 1D vectorial descriptors $\{v_1, v_2, ..., v_n, ..., v_N\}$, where $v_n\in\mathbb{R}^{L}$. The $\phi_n(.)$ corresponding to $x_n$ is defined as:

\begin{align}
    &v_n = \phi_n(x_n) = f^{\rm max\_pool}_{H\times W}(x_n''),\\
    &x_n''=f^{\rm ReLU}(f^{\rm bn}(f^{\rm conv}_{3\times3\times \frac{L}{2} \times L}(x_n'))),\\
    &x_n'=f^{\rm ReLU}(f^{\rm bn}(f^{\rm conv}_{1\times1\times C \times \frac{L}{2}}(x_n))),
\end{align}
where $f^{\rm max\_pool}_{H\times W}(.)$ denotes a max-pooling operation whose window size is $H\times W$. $f^{\rm conv}(.)$ illustrates the 2D convolution operation by kernel size. For example, $f^{\rm conv}_{1\times1\times C \times \frac{L}{2}}(.)$ denotes a 2D convolution operation whose kernel size is $1\times1\times C \times \frac{L}{2}$ ($1\times1$ is the spatial size, $C$ is the number of input channels, and $\frac{L}{2}$ is the number of output channels). $f^{\rm bn}(.)$ denotes the batch normalization operation~\cite{ioffe2015batch}, and $f^{\rm ReLU}(.)$ denotes the ReLU operation. 

Thereafter, we use a set of operations $\{\psi_1(.)$, $\psi_2(.)$, ..., $\psi_n(.)$, ..., $\psi_N(.)\}$ to respectively process $\{v_1$, $v_2$, ..., $v_n$, ..., $v_N\}$ to predict the probability distribution $\{p_1$, $p_2$, ..., $p_n$, ..., $p_N\}$ over the classes at each stage:

\begin{align}
\begin{split}
    p_n &= \psi_n(v_n)\\
    &=f^{\rm fc}_{\frac{L}{2}\times K}(f^{\rm ReLU}(f^{\rm bn}(f^{\rm fc}_{L\times\frac{L}{2}}(f^{\rm bn}(v_n))))),
\end{split}
\end{align}
where $p_n\in\mathbb{R}^{K}$, and $K$ denotes the number of the classes of driving behaviors. $f^{\rm fc}_{\frac{L}{2}\times K}(.)$ denotes a fully connected layer whose input size is $\frac{L}{2}$ and the output size is $K$. $f^{\rm fc}_{L\times \frac{L}{2}}(.)$ denotes a fully connected layer whose input size is $L$ and the output size is $\frac{L}{2}$.

After the last stage $s_n$, we add an additional stage by concatenating $v_1$, $v_2$, ..., $v_N$ and generating the concatenated vector into the probability distribution over the classes as:
\begin{align}
\begin{split}
    p_{N+1} &= \psi_{N+1}(v_{N+1})\\
    &=f^{\rm fc}_{\frac{L}{2}\times K}(f^{\rm ReLU}(f^{\rm bn}(f^{\rm fc}_{NL\times\frac{L}{2}}(f^{\rm bn}(v_{N+1}))))),
\end{split}\\
v_{N+1} &= f^{\rm concat}(v_1,v_2, ..., v_n, ..., v_N),
\end{align}
where $f^{\rm concat}(.)$ denotes the concatenation operation. Now, we have $N+1$ prediction probability distributions $\{p_1$, $p_2$, ..., $p_n$, ..., $p_N$, $p_{N+1}\}$. The teacher network is trained by using a cross entropy loss $\mathcal{L}_{\rm cls}(.)$ to minimize the distance between ground truth label $p_{\rm truth}$ and each prediction probability distribution of $\{p_1$, $p_2$, ..., $p_n$, ..., $p_N$, $p_{N+1}\}$:

\begin{align}
\mathcal{L}_{\rm cls}(p_n,p_{\rm truth}) = -\sum^{K}_{k=1}p_{\rm truth}^{(k)} \log(p_{n}^{(k)}),
\end{align}
where $p_{n}^{(k)}$ denotes the probability that the input belongs to the category $k$ at the stage $s_n$. $p_{\rm truth}^{(k)}$ equals to 1 if it is true that the input belongs to the category $k$, and equals to 0 on the contrary.

The overall algorithm of building  the  teacher  network is given in Algorithm~\ref{alg:algorithm1}. For the stages $s_1\sim s_n$, the input images are augmented with Imgaug~\cite{imgaug}.

\subsection{Distillation-Based Neural Architecture Search for the Student Network}
The computation overhead, including speed and parameter size, acts as an extremely crucial role for DDR. According to the experiences of previous studies~\cite{qin2021distracted,baheti2018detection}, it is much more favorable to use large convolution filters rather than deep layers because the former is able to compute in parallel to achieve a fast processing speed that satisfies the requirements of the real-world application. Thus, in this work, we design the student network to have four convolutional blocks, which are followed by a global average pooling layer (GAP) and a fully connected (FC) layer for predicting the probability distribution over the classes.

\begin{figure*}[t!]
\centering\includegraphics[width=\linewidth]{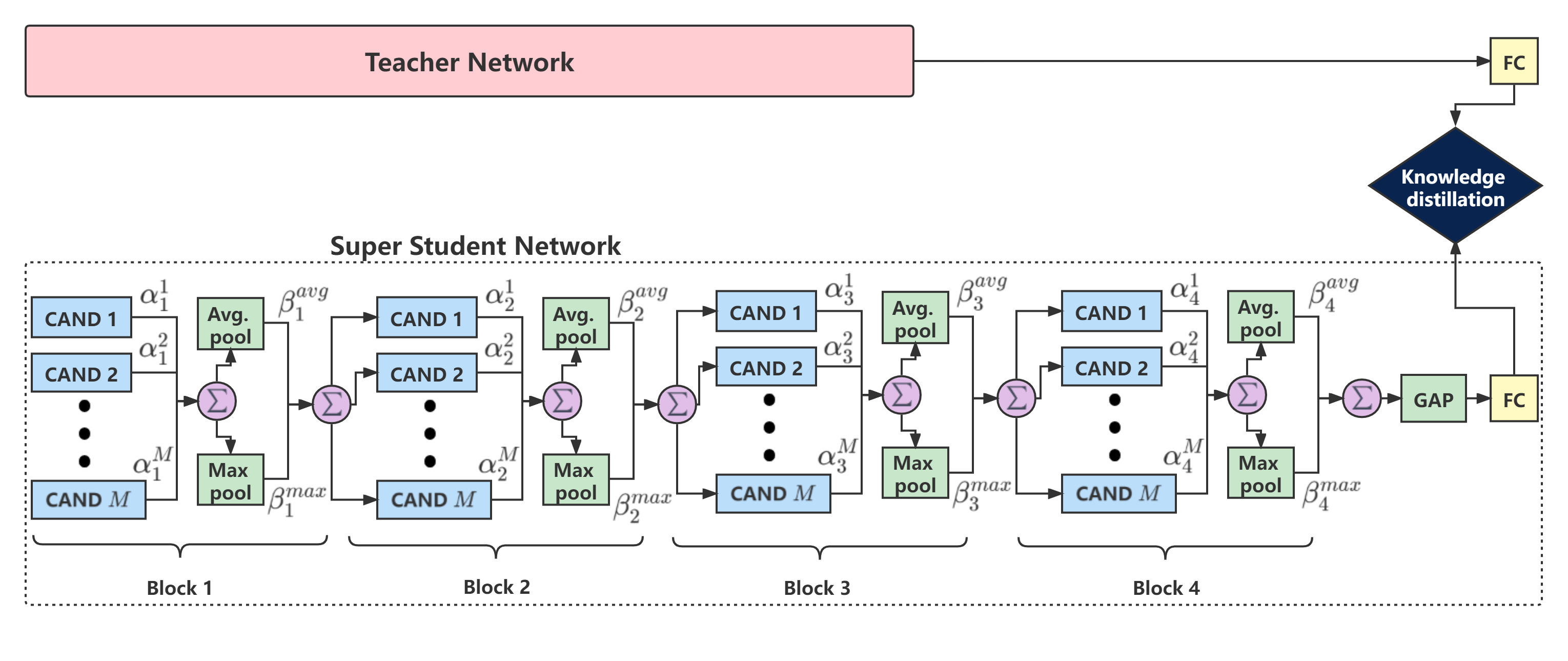}
\caption{Illustration of the searching process. ``CAND'' is the abbreviation for ``candidate''.  In the super student network, there are several candidates for the convolutional architectures of each block.  Besides, there are two candidates of the pooling method, namely average pooling and max pooling, in each block. The candidates of convolutional architecture and pooling methods for each block are aggregated by the weighted sum. $\alpha$ and $\beta$ are the learnable weights. ``GAP'' means global average pooling and ``FC'' means the fully-connected operation. The super student network is trained to learn from the teacher network by knowledge distillation. After the training, only the candidates with the maximum weight are kept and forms the student network.}
\label{fig:ret}
\end{figure*}

For each block, we use pyramidal convolution (PyConv)~\cite{duta2020pyramidal} rather than standard convolution. PyConv contains a pyramid of kernels, where each level involves different types of filters with varying sizes. Using PyConv for DDR has two benefits. First, PyConv can capture different levels of details in the scene. A filter of a smaller kernel size has smaller receptive fields and thus can capture more local information and more detailed clues. A filter of a bigger kernel size has bigger receptive fields and thus can ``see'' more information at once and capture relatively more global information, such as the dependencies among some local patterns, some large objects, etc. Such multi-level details are very important for recognizing driver behaviors. Second, PyConv is flexible and extensible, giving a large space of potential architecture designs. That is, PyConv gives strong potential to search for a lightweight architecture. 

At the end of each block, we use a pooling layer to downsample the feature maps. Two types of pooling layers are widely used for this objective: max pooling and average pooling. We define our search space as the candidates of different designs of PyConv and different pooling types in the four convolutional blocks.

As shown in Figure~\ref{fig:ret}, the overall process of defining the architecture of the student network is given as: at first, we construct a super student network covering all the candidates of each block. In the super student network, the output feature maps of the candidates of each block are aggregated by a weighted sum to become the input of the next block. The sum weights are learnable and represent the probability of choosing the candidates. Then the super student network is trained to learn from the teacher network. Thereafter, the final architecture of the student network is derived by selecting the candidate with the maximum probability.

Specifically, let $\{b_1, b_2, b_3, b_4\}$ denote the four blocks of the student network and super student network. $\{\omega^1_{b}(.), \omega^2_{b}(.), ..., \omega^m_b(.), ..., \omega_b^M(.)\}$ denote $M$ different candidates of PyConv for the block $b$. $\{f^{\rm avg\_pool}_b(.), f^{\rm max\_pool}_b(.)\}$ denotes the candidates of using average pooling or max pooling layer at the end of the block $b$. Given the feature map $X^{\rm in}_b$ outputted by the previous block, the output feature map $X^{\rm out}_b$ of the block $b$ in the super student network is defined as:

\begin{align}
\begin{split}
X^{\rm out}_b = &\beta_b^{\rm avg} f^{\rm avg\_pool}_b(\sum_{m=1}^M \alpha^m_b \omega^m_{b} (X^{\rm in}_b))\\
&+ \beta_b^{\rm max} f^{\rm max\_pool}_b(\sum_{m=1}^M \alpha^m_b \omega^m_{b} (X^{\rm in}_b)),
\end{split}
\end{align}
where $\{\alpha^1_{b}, \alpha^2_{b}, ..., \alpha^m_b, ..., \alpha_b^M\}$ and $\{\beta_b^{\rm avg}, \beta_b^{\rm max}\}$ are the probabilities of choosing the corresponding candidates, and they are computed as:

\begin{align}
 &\alpha^m_{b} = \frac{\exp(\hat{\alpha}^m_{b})}{\sum^{M}_{j=1} \exp(\hat{\alpha}^j_{b})},\\
  &\beta^g_{b} = \frac{\exp(\hat{\beta}^g_{b})}{\exp(\hat{\beta}^{\rm avg}_{b})+\exp(\hat{\beta}^{\rm max}_{b})},~ g=\{\rm avg,max\},
\end{align}
where, $\{\hat{\alpha}^1_{b}, \hat{\alpha}^2_{b}, ..., \hat{\alpha}^m_b, ..., \hat{\alpha}_b^M\}$ and $\{\hat{\beta_b}^{\rm avg}, \hat{\beta_b}^{\rm max}\}$ are learnable parameters that are all initialized as 1 and optimized during the training.

All the blocks of $\{b_1, b_2, b_3, b_4\}$ of the super student network are constructed by the process introduced above. The output feature map of $b_4$ is processed by the GAP and FC layers to predict the probability distribution over the classes ($p_{\rm super}$). As mentioned above, after the training of the teacher network, we only use $p_{N+1}$ of the teacher network for category prediction. The super student network is trained with the search loss $\mathcal{L}_{\rm search}(.)$ defined as:
\begin{align}
\label{equ:loss1}
\begin{split}
&\mathcal{L}_{\rm \rm search}(p_{\rm super}, p_{N+1}, p_{\rm truth}) =\\ &\lambda \mathcal{L}_{\rm mse}(p_{\rm super},p_{N+1}) + (1-\lambda)\mathcal{L}_{\rm cls}(p_{\rm super},p_{\rm truth}),
\end{split}
\end{align}
where ${L}_{\rm mse}(.)$ denotes the mean squared error loss, and $\lambda$ is a manual hyperparameter. During the training of the super student network, the parameters of the teacher network are fixed. After the training, we only keep the candidate with the maximum probability and prune all the other candidates for each block to construct the student network.

\begin{figure*}[t!]
\centering\includegraphics[width=\linewidth]{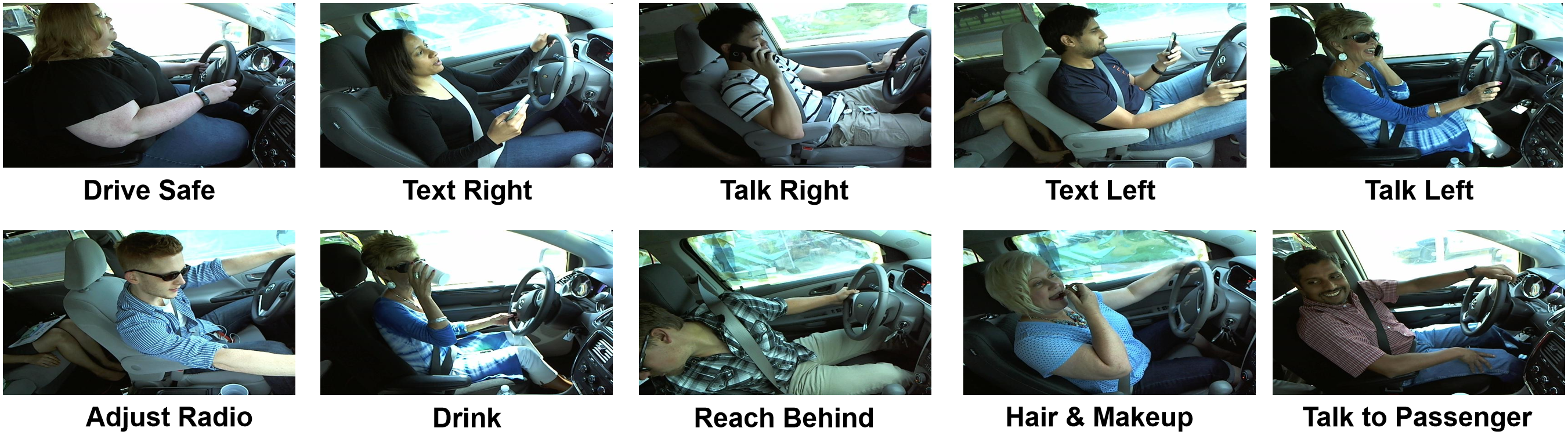}
\caption{Sample images of distracted driving behaviors on the Statefarm Distracted Driver Detection Dataset (SFD3)~\cite{statefarm}.}
\label{fig:sfd3}
\end{figure*}

\begin{figure*}[ht!]
\centering\includegraphics[width=\linewidth]{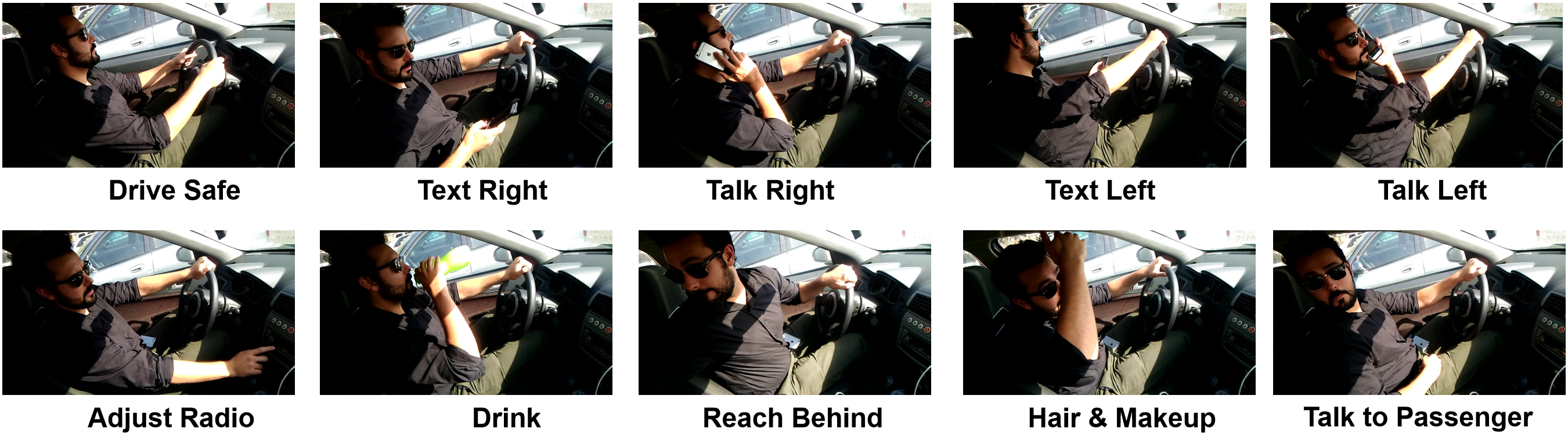}
\caption{Sample images of distracted driving behaviors on the AUC Distracted Driver Dataset (AUCD2)~\cite{alotaibi2019distracted}.}
\label{fig:aucd2}
\end{figure*}

\begin{figure*}[t!]
\centering\includegraphics[width=\linewidth]{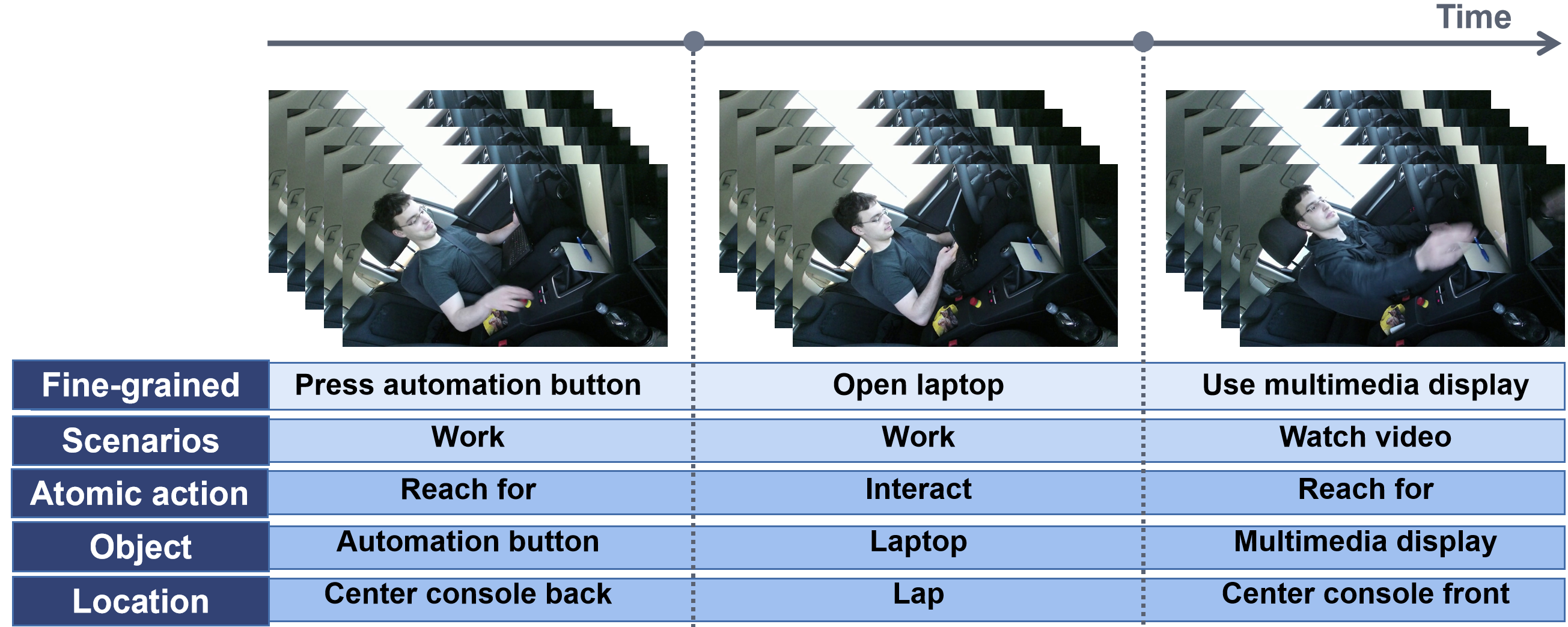}
\caption{Sample frames of distracted driving behaviors on the Drive\&Act Dataset (DAD)~\cite{drive_and_act_2019_iccv}. The video clips are labeled with multiple annotations, including different categories of fine-grained activities, scenarios, atomic actions, objects, locations, together with possible combinations of the atomic actions, objects, and locations.}
\label{fig:drive_act}
\end{figure*}

\subsection{Knowledge Transfer}
In the former subsection, we use the teacher network to guide the search of the student network architecture, and in this subsection, we use it to transfer knowledge to the student network. Assume that $p_{\rm student}$ is the probability distribution over the classes predicted by the student network. The student network is trained with the knowledge transfer loss $\mathcal{L}_{\rm trans}(.)$ defined as:

\begin{align}
\label{equ:loss2}
\begin{split}
&\mathcal{L}_{\rm trans}(p_{\rm student}, p_{N+1}, p_{\rm truth}) =\\ &\lambda \mathcal{L}_{\rm mse}(p_{\rm student},p_{N+1}) + (1-\lambda)\mathcal{L}_{\rm cls}(p_{\rm student},p_{\rm truth}).
\end{split}
\end{align}

\section{Datasets and Implementation Details}
\subsection{Dataset Description}

The experiments are conducted using two types of datasets: image-based DDR dataset and video-based DDR dataset. The image-based DDR task requires recognizing the driver's behavior from each given image. The video-based DDR task requires recognizing the driver's behavior from each given video clip containing several frames. We mainly carried out the experiments of training the teacher network, defining the student network, and evaluating the performance of the teacher and student networks on the image-based DDR datasets. Then, we obtained an extremely lightweight yet powerful student network for the image-based DDR task. Thereafter, following Hara \textit{et al.}~\cite{hara2018can}, we extended the student network from 2D to 3D for the video-based DDR task.

For the image-based DDR task, we carried out experiments on two standard benchmark datasets for DDR: the Statefarm Distracted Driver Detection Dataset (SFD3)~\cite{statefarm} and the AUC Distracted Driver Dataset (AUCD2)~\cite{alotaibi2019distracted}. These two datasets are the most widely used datasets, and have been used for many studies on DDR. Both of the two datasets are composed of one safe driving action and nine distracted driving actions including (\romannumeral1) text right, (\romannumeral2) talk right, (\romannumeral3) text left, (\romannumeral4) talk left, (\romannumeral5) adjust radio, (\romannumeral6) drink, (\romannumeral7) reach behind, (\romannumeral8) hair and makeup, and (\romannumeral9) talk to passenger. The images of both datasets are taken by dashboard cameras recording the driver's behavior. The sample images of the SFD3 and AUCD2 are shown in Figure~\ref{fig:sfd3} and Figure~\ref{fig:aucd2}, respectively.

SFD3 is one of the most influential public datasets in the field of DDR. There are 22,424 images for training (around 2,000 images in each category) and 79,728 unlabeled images for testing. Since SFD3 does not provide the labels for the testing images, we follow the common practice of previous studies to perform experiments on the training dataset. We randomly split the training dataset of SFD3 as training image: testing image = 7:3~\cite{dhakate2020distracted,qin2021distracted}, 7.5:2.5~\cite{masood2020detecting,hssayeni2017distracted,qin2021distracted}, 8:2~\cite{majdi2018drive,janet2020real,huang2020hcf,qin2021distracted,zhang_2016}, 9:1~\cite{okon2017detecting,qin2021distracted}. In this work, for each proportion of the train-test partition, we randomly split the images 10 times and report the average accuracy.

AUCD2 is another widely used public dataset for DDR. It has 17,308 RGB frames, of which 12,977 are for training, while the remaining 4,331 are for testing. 

For the video-based DDR task, we utilized the Drive\&Act Dataset (DAD)~\cite{drive_and_act_2019_iccv}. This is a large-scale video dataset consisting of various driver activities, with more than 9.6 million frames. As shown in Figure~\ref{fig:drive_act}, the DAD provides multiple annotations for performing three types of recognition tasks on the video clips. The first task is the scenario recognition task, which requires recognizing the top-level activities (e.g., work and drink) from each given video clip. There are totally 12 different scenario categories. The second task is the fine-grained activity recognition task, which requires recognizing the specific semantic actions (e.g., open laptop, close bottle, etc.) from each video clip. There are totally 34 different categories of fine-grained activities. The third task is the atomic action unit recognition task. The atomic action units portray the lowest degree of abstraction and are basic driver interactions with the environment. The annotations of the atomic action units involve triplets of atomic action, object, and location, which are detached from long-term semantic meaning and can be regarded as building blocks for complex activities. There are five categories of atomic actions (e.g., reach for), 17 categories of objects (e.g., automation button), 14 categories of locations (e.g., center console back), and 372 possible combinations.

\begin{table*}[t!]
\centering
\caption{Candidates of each block}\label{tab:candidates}
\begin{tabular}{l|ll|ll|ll|ll}
\hline
Blocks &
  \multicolumn{2}{l|}{Block 1 ($b_1$)} &
  \multicolumn{2}{l|}{Block 2 ($b_2$)} &
  \multicolumn{2}{l|}{Block 3 ($b_3$)} &
  \multicolumn{2}{l}{Block 4 ($b_4$)} \\  \hline
Output Size &
  \multicolumn{2}{l|}{112$\times$112, 32} &
  \multicolumn{2}{l|}{56$\times$56, 64} &
  \multicolumn{2}{l|}{28$\times$28, 128} &
  \multicolumn{2}{l}{14$\times$14, 256} \\  \hline
Candidate 1 &
  \begin{tabular}[c]{@{}l@{}}$\begin{bmatrix} 11\times11,16, 1\\7\times7,16,  1\end{bmatrix}$\end{tabular} &
  \begin{tabular}[c]{@{}l@{}}Avg. Pool\\ $\begin{bmatrix}2\times2\end{bmatrix}$\end{tabular} &
  \begin{tabular}[c]{@{}l@{}}$\begin{bmatrix} 9\times9,32, 1\\5\times5,32,  1\end{bmatrix}$\end{tabular} &
  \begin{tabular}[c]{@{}l@{}}Avg. Pool\\ $\begin{bmatrix}2\times2\end{bmatrix}$\end{tabular} &
  \begin{tabular}[c]{@{}l@{}}$\begin{bmatrix} 5\times5,64, 1\\3\times3,64,  1\end{bmatrix}$\end{tabular} &
  \begin{tabular}[c]{@{}l@{}}Avg. Pool\\ $\begin{bmatrix}2\times2\end{bmatrix}$\end{tabular} &
  \begin{tabular}[c]{@{}l@{}}$\begin{bmatrix} 3\times3,128, 1\\1\times1,128,  1\end{bmatrix}$\end{tabular} &
  \begin{tabular}[c]{@{}l@{}}Avg. Pool\\ $\begin{bmatrix}2\times2\end{bmatrix}$\end{tabular} \\ 
Candidate 2 &
  \begin{tabular}[c]{@{}l@{}}$\begin{bmatrix} 11\times11,16, 1\\5\times5,16,  1\end{bmatrix}$\end{tabular} &
  \begin{tabular}[c]{@{}l@{}}Max Pool\\ $\begin{bmatrix}2\times2\end{bmatrix}$\end{tabular} &
$\begin{bmatrix} 9\times9,32, 1\\5\times5,32,  2\end{bmatrix}$   &
  \begin{tabular}[c]{@{}l@{}}Max Pool\\ $\begin{bmatrix}2\times2\end{bmatrix}$\end{tabular} &
$\begin{bmatrix} 5\times5,64, 1\\3\times3,64,  2\end{bmatrix}$   &
  \begin{tabular}[c]{@{}l@{}}Max Pool\\ $\begin{bmatrix}2\times2\end{bmatrix}$\end{tabular} &
$\begin{bmatrix} 3\times3,128, 1\\1\times1,128,  2\end{bmatrix}$   &
  \begin{tabular}[c]{@{}l@{}}Max Pool\\ $\begin{bmatrix}2\times2\end{bmatrix}$\end{tabular} \\ 
Candidate 3 &
  \begin{tabular}[c]{@{}l@{}}$\begin{bmatrix} 11\times11,16, 1\\3\times3,16,  1\end{bmatrix}$\end{tabular} &
  - &
$\begin{bmatrix} 9\times9,16, 1\\5\times5,16,  4\end{bmatrix}$   &
  - &
$\begin{bmatrix} 5\times5,64, 1\\3\times3,64,  4\end{bmatrix}$   &
  - &
$\begin{bmatrix} 3\times3,128, 1\\1\times1,128,  4\end{bmatrix}$   &
  - \\ 
Candidate 4 &
  \begin{tabular}[c]{@{}l@{}}$\begin{bmatrix} 11\times11,16, 1\\1\times1,16,  1\end{bmatrix}$\end{tabular} &
  - &
$\begin{bmatrix} 9\times9,32, 1\\3\times3,32,  1\end{bmatrix}$   &
  - &
$\begin{bmatrix} 5\times5,64, 1\\1\times1,64,  1\end{bmatrix}$   &
  - &
  - &
  - \\ 
Candidate 5 &
  \begin{tabular}[c]{@{}l@{}}-\end{tabular} &
  - &
$\begin{bmatrix} 9\times9,32, 1\\3\times3,32,  2\end{bmatrix}$   &
  - &
$\begin{bmatrix} 5\times5,64, 1\\1\times1,64,  2\end{bmatrix}$   &
  - &
-   &
  - \\ 
Candidate 6 &
  - &
  - &
$\begin{bmatrix} 9\times9,32, 1\\3\times3,32,  4\end{bmatrix}$   &
  - &
$\begin{bmatrix} 5\times5,64, 1\\1\times1,64,  4\end{bmatrix}$   &
  - &
-   &
  - \\ 
Candidate 7 &
  - &
  - &
$\begin{bmatrix} 9\times9,32, 1\\1\times1,32,  1\end{bmatrix}$   &
  - &
-   &
  - &
-   &
  - \\ 
Candidate 8 &
  - &
  - &
$\begin{bmatrix} 9\times9,32, 1\\1\times1,32,  2\end{bmatrix}$   &
  - &
-   &
  - &
-   &
  - \\
Candidate 9 &
  - &
  - &
$\begin{bmatrix} 9\times9,32, 1\\1\times1,32,  4\end{bmatrix}$   &
  - &
-   &
  - &
-   &
  - \\ \hline
\end{tabular}
\end{table*}

\begin{table*}[t!]
\caption{Probabilities of choosing each candidate of each block} \label{tab:candidate_prob}
\centering
\setlength{\tabcolsep}{6mm}
\begin{tabular}{l|ll|ll|ll|ll}
\hline
 &
  \multicolumn{2}{l|}{Block 1 ($b_1$)} &
  \multicolumn{2}{l|}{Block 2 ($b_2$)} &
  \multicolumn{2}{l|}{Block 3 ($b_3$)} &
  \multicolumn{2}{l}{Block 4 ($b_4$)} \\ \hline
Candidate 1 &
  0.228 &
  0.322 &
  \cellcolor[HTML]{C0C0C0}0.139 &
  0.201 &
  \cellcolor[HTML]{C0C0C0}0.2074 &
  0.043 &
  \cellcolor[HTML]{C0C0C0}0.966 &
  0.052 \\ 
Candidate 2 &
  0.246 &
  \cellcolor[HTML]{C0C0C0}0.678 &
  0.121 &
  \cellcolor[HTML]{C0C0C0}0.799 &
  0.1865 &
  \cellcolor[HTML]{C0C0C0}0.957 &
  0.018 &
  \cellcolor[HTML]{C0C0C0}0.948 \\ 
Candidate 3 & \cellcolor[HTML]{C0C0C0}0.264 & - & 0.127 & - & 0.1622 & - & 0.016 & - \\ 
Candidate 4 & 0.262                         & - & 0.123 & - & 0.1504 & - & -     & - \\ 
Candidate 5 & -                             & - & 0.101 & - & 0.1507 & - & -     & - \\
Candidate 6 & -                             & - & 0.105 & - & 0.1428 & - & -     & - \\
Candidate 7 & -                             & - & 0.097 & - & -      & - & -     & - \\
Candidate 8 & -                             & - & 0.096 & - & -      & - & -     & - \\
Candidate 9 & -                             & - & 0.091 & - & -      & - & -     & - \\ \hline
\end{tabular}
\end{table*}

\begin{table*}[t!]
\caption{The student network defined by distillation-based neural architecture search} \label{tab:candidate_results}

\centering
\setlength{\tabcolsep}{7mm}
\begin{threeparttable}
\begin{tabular}{l|l|l|l|l}
\hline
Block 1 ($b_1$) &
  Block 2 ($b_2$) &
  Block 3 ($b_3$) &
  Block 4 ($b_4$) &
  \multirow{2}{*}{\begin{tabular}[c]{@{}l@{}}GAP\\ $\mathscr{N}$-D FC\!\\ Softmax\!\end{tabular}} \\ \cline{1-4}
\begin{tabular}[c]{@{}l@{}}{\!\!\fontsize{9pt}{0pt} $\begin{bmatrix} 11\times11,16, 1\\3\times3,16,  1\end{bmatrix}$}\!\\ Batch Norm.\\ ReLU\\ Max Pool\\$\begin{bmatrix}2\times2\end{bmatrix}$\end{tabular} &
  \begin{tabular}[c]{@{}l@{}}{\!\!\fontsize{9pt}{0pt} $\begin{bmatrix} 9\times9,32, 1\\5\times5,32,  1\end{bmatrix}$}\!\\ Batch Norm.\\ ReLU\\ Max Pool\\$\begin{bmatrix}2\times2\end{bmatrix}$\end{tabular} &
  \begin{tabular}[c]{@{}l@{}}{\!\!\fontsize{9pt}{1pt} $\begin{bmatrix} 5\times5,64, 1\\3\times3,64,  1\end{bmatrix}$}\!\\ Batch Norm.\\ ReLU\\ Max Pool\\$\begin{bmatrix}2\times2\end{bmatrix}$\end{tabular} &
  \begin{tabular}[c]{@{}l@{}}{\!\!\fontsize{9pt}{1pt} $\begin{bmatrix} 3\times3,128, 1\\1\times1,128,  1\end{bmatrix}$}\!\\ Batch Norm.\\ ReLU\\ Max Pool\\$\begin{bmatrix}2\times2\end{bmatrix}$\end{tabular} &
   \\ \hline
\begin{tabular}[c]{@{}l@{}}112$\times$112, 32\end{tabular} &
56$\times$56, 64   &
28$\times$28, 128   &
14$\times$14, 256    &
1$\times$1, $\mathscr{N}$   \!\\ \hline
\end{tabular}
\begin{tablenotes}
        \item[*] $\mathscr{N}$ denotes the number of categories.
\end{tablenotes}
\end{threeparttable}
\end{table*}

\subsection{Implementation Details}
\textbf{Teacher Network.} We use SKResNeXt50\_32$\times$4d~\cite{li2019selective} as the backbone of the feature extractor $E$. We divide SKResNeXt50\_32$\times$4d into three segments $\{m_1,m_2,m_3\}$. $m_1$ includes the Conv1--Conv3 of SKResNeXt50\_32$\times$4d. $m_2$ and $m_3$ respectively include the Conv4 and Conv5 of SKResNeXt50\_32$\times$4d.

\textbf{Super Student Network.} As mentioned above, we define our search space as the candidates of different designs for the four convolution blocks of the student network and construct a super student network to cover all the candidates. The specific candidates are shown in Table~\ref{tab:candidates}. In Table~\ref{tab:candidates}, the design of filters are illustrated by kernel size, number of channels, and number of groups. For example, $\begin{bmatrix} 11\times11,16, 1\\7\times7,16,  1\end{bmatrix}$ denotes a PyConv layer with two types of filters: one filter has $11\times11$ kernel size and the other has $7\times7$ kernel size. Both filters have 16 channels and 1 group. The pooling layers are illustrated by the type and window size. For example, ``Avg. Pool $\begin{bmatrix}2\times2\end{bmatrix}$'' denotes an average pooling layer with $2\times2$ window size. The stride of all the convolution layers is set as 1 and the padding size is set as $\frac{\theta-1}{2}$, where $\theta$ is the spatial size of the filter. Thus, the convolution layers do not change the spatial size of feature maps. The stride of the pooling layers is set as 2, and the height and width of feature maps decrease by half after the pooling layers.

\textbf{Training Details.} For the experiments on the image-based datasets, during the training, all the learning rate are set as 0.002 with cosine annealing~\cite{cosineannealing}. Weight decay is set as $5\times10^{-4}$. The input images are resized to $256\times256$ and applied with random crop of $224\times224$ region for training, center crop of $224\times224$ region for testing. We set batch size as 32 and train each network for 300 epochs. The manual hyper parameter $\lambda$ in Equation~\ref{equ:loss1} and Equation~\ref{equ:loss2} is set as 0.7, which is a common setting for distillation. For the experiments on the video-based dataset, we follow the settings of Hara \textit{et al.}~\cite{hara2018can}. Specifically, the learning rate is set as 0.001 with plateau scheduler~\cite{hara2018can}. Weight decay is set as $1\times10^{-5}$. The batch size is set as 32, and 16 frames ($16\times3\times112\times112$)  are sampled for each video clip by uniform sampling.

\begin{table*}[t!]
\setlength{\tabcolsep}{6mm}
\caption{The recognition accuracy of the teacher and student networks} \label{tab:performance}
\centering
\begin{threeparttable}
\begin{tabular}{lllll}
\hline
                       & \multicolumn{2}{l}{Teacher Network}                        & \multicolumn{2}{l}{Student Network}   \\ \hline
 &
  \multicolumn{1}{l}{Backbone} &
  \multicolumn{1}{l}{\begin{tabular}[c]{@{}l@{}}Backbone +PL\end{tabular}} &
  \multicolumn{1}{l}{\begin{tabular}[c]{@{}l@{}}Train from Scratch\end{tabular}} &
  \begin{tabular}[c]{@{}l@{}}Finetune after  Distillation\end{tabular} \\ \hline
AUCD2                  & \multicolumn{1}{l}{\fontsize{9pt}{0pt} {95.29\%}} & \multicolumn{1}{l}{\fontsize{9pt}{0pt} {96.35\%}} & \multicolumn{1}{l}{\fontsize{9pt}{0pt} {95.12\%}} & \fontsize{9pt}{0pt} {95.64\%} \\ \hline
                       & \multicolumn{4}{c}{Train:Test=7:3}                                           \\ 
                       & \multicolumn{1}{l}{\fontsize{9pt}{0pt} {99.75$\pm$0.12\%}} & \multicolumn{1}{l}{\fontsize{9pt}{0pt} {99.87$\pm$0.03\%}} & \multicolumn{1}{l}{\fontsize{9pt}{0pt} {99.81$\pm$0.09\%}} & \fontsize{9pt}{0pt} {99.87$\pm$0.03\%} \\ \cline{2-5} 
                       & \multicolumn{4}{c}{Train:Test=7.5:2.5}                                      \\
                       & \multicolumn{1}{l}{\fontsize{9pt}{0pt} {99.79$\pm$0.11\%}} & \multicolumn{1}{l}{\fontsize{9pt}{0pt} {99.88$\pm$0.04\%}} & \multicolumn{1}{l}{\fontsize{9pt}{0pt} {99.82$\pm$0.09\%}} & \fontsize{9pt}{0pt} {99.88$\pm$0.04\%} \\ \cline{2-5} 
                       & \multicolumn{4}{c}{Train:Test=8:2}                                          \\ 
                       & \multicolumn{1}{l}{\fontsize{9pt}{0pt} {99.82$\pm$0.09\%}} & \multicolumn{1}{l}{\fontsize{9pt}{0pt} {99.89$\pm$0.05\%}} & \multicolumn{1}{l}{\fontsize{9pt}{0pt} {99.87$\pm$0.07\%}} & \fontsize{9pt}{0pt} {99.89$\pm$0.05\%} \\ \cline{2-5} 
                       & \multicolumn{4}{c}{Train:Test=9:1}                                          \\ 
\multirow{-8}{*}{SFD3} & \multicolumn{1}{l}{\fontsize{9pt}{0pt} {99.87$\pm$0.05\%}} & \multicolumn{1}{l}{\fontsize{9pt}{0pt} {99.91$\pm$0.05\%}} & \multicolumn{1}{l}{\fontsize{9pt}{0pt} {99.87$\pm$0.05\%}} & \fontsize{9pt}{0pt} {99.91$\pm$0.05\%} \\ \hline
\end{tabular}
\begin{tablenotes}
        \item[*] For SFD3, we illustrate the accuracy range of 10 random splits.
\end{tablenotes}
\end{threeparttable}
\end{table*}

\begin{table}[t!]
\setlength{\tabcolsep}{2mm}
\caption{The F1-score of the teacher and student networks} \label{tab:F_1_score}
\begin{tabular}{lllll}
\hline
                  & \multicolumn{2}{l}{Teacher Network} & \multicolumn{2}{l}{Student Network} \\ \hline
 &
  Backbone &
  \begin{tabular}[c]{@{}l@{}}Backbone\\ +PL\end{tabular} &
  \begin{tabular}[c]{@{}l@{}}Train from\\ Scratch\end{tabular} &
  \begin{tabular}[c]{@{}l@{}}Finetune \\ after  \\ Distillation\end{tabular} \\ \hline
Safe driving      & 93.60\%          & 95.14\%          & 93.43\%          & 94.07\%          \\
Text Right        & 95.24\%          & 96.44\%          & 95.13\%          & 95.73\%          \\
Talk Right        & 94.74\%          & 96.21\%          & 94.25\%          & 94.93\%          \\
Text left         & 95.09\%          & 96.04\%          & 95.09\%          & 95.40\%          \\
Talk left         & 95.86\%          & 97.35\%          & 95.42\%          & 95.86\%          \\
Adjust Radio      & 95.58\%          & 96.56\%          & 95.42\%          & 95.60\%          \\
Drink             & 96.54\%          & 97.29\%          & 96.30\%          & 97.04\%          \\
Reach Behind      & 95.41\%          & 96.30\%          & 95.41\%          & 95.76\%          \\
Hair \& Makeup    & 95.64\%          & 96.48\%          & 95.64\%          & 96.48\%          \\
Talk to Passenger & 96.68\%          & 97.00\%          & 96.60\%          & 96.84\%          \\ \hline
Average           & 95.44\%          & 96.48\%          & 95.27\%          & 95.77\%          \\ \hline
\end{tabular}
\end{table}

\begin{table}[t!]
\setlength{\tabcolsep}{1.7mm}
\caption{Comparison of the designed student networks with existing lightweight networks in terms of GFLOPs and time consumption}\label{tab:GFLOPS_TIME_COST}
\begin{tabular}{lllll}
\hline
 &                    & GFLOPs & \multicolumn{2}{l}{Time Cost} \\ \hline
\multirow{7}{*}{Image} &
   &
   &
  Single Image &
  \begin{tabular}[c]{@{}l@{}}A Batch of \\ 32 Images\end{tabular} \\
 & MobileVGG~\cite{baheti2020towards}          & 2.11   & 5.19ms        & 122.34ms      \\
 & MobileNet~\cite{howard2017mobilenets}          & 0.59   & 3.54ms        & 53.15ms       \\
 & MobileNetV2~\cite{8578572}        & 0.33   & 6.94ms        & 44.72ms       \\
 & SqueezeNet~\cite{iandola2016squeezenet}    & 0.86   & 3.86ms        & 48.79ms       \\
 & D-HCNN~\cite{qin2021distracted}   & 31.10  & 7.40ms        & 40.67ms       \\
 & \textbf{2D Student Network} & \textbf{2.25}   & \textbf{2.23ms}        & \textbf{35.69ms}       \\ \hline
\multirow{5}{*}{\begin{tabular}[c]{@{}l@{}}Video\\ Clip\end{tabular}} &
   &
   &
  \begin{tabular}[c]{@{}l@{}}Single Video\\ Clip\end{tabular} &
  \begin{tabular}[c]{@{}l@{}}A Batch of\\ 8 Clips\end{tabular} \\
 & C3D~\cite{tran2015learning}    & 38.55  & 25.57ms       & 1452.86ms     \\
 & P3D ResNet~\cite{qiu2017learning}  & 18.67  & 148.04ms      & 983.22ms      \\
 & I3D~\cite{carreira2017quo}      & 27.90  & 61.86ms       & 588.44ms      \\
 & \textbf{3D Student Network} & \textbf{37.20}  & \textbf{25.35ms}       & \textbf{479.83ms}      \\ \hline
\end{tabular}
\end{table}

\begin{table}[t!]
\begin{center}
\caption{Comparison with state-of-the-art methods on AUCD2} \label{tab:soa:aucd2}
\begin{tabular}{lll}
\hline
\multicolumn{1}{l}{\textbf{Approach}} & \multicolumn{1}{l}{\textbf{\begin{tabular}[c]{@{}l@{}}Parameter \\ Size\end{tabular}}} & \textbf{Accuracy} \\ \hline
AlexNet on Original Scene~\cite{abouelnaga2017real}                                        & 62M                                  & 93.65\%          \\
AlexNet on Skin Segmentation~\cite{abouelnaga2017real}                                     & 62M                                  & 93.60\%          \\
AlexNet on Face Segmentation~\cite{abouelnaga2017real}                                     & 62M                                  & 86.68\%          \\
AlexNet on Hand Segmentation~\cite{abouelnaga2017real}                                     & 62M                                  & 89.52\%          \\
AlexNet on Face + Hand Segmentation~\cite{abouelnaga2017real}                              & 62M                                  & 86.68\%          \\
AlexNet on Original Scene~\cite{abouelnaga2017real}                                        & 24M                                  & 95.17\%          \\
AlexNet on Skin Segmentation~\cite{abouelnaga2017real}                                     & 24M                                  & 94.57\%          \\
AlexNet on Face Segmentation~\cite{abouelnaga2017real}                                     & 24M                                  & 88.82\%          \\
AlexNet on Hand Segmentation~\cite{abouelnaga2017real}                                     & 24M                                  & 91.62\%          \\
AlexNet on Face + Hand Segmentation~\cite{abouelnaga2017real}                              & 24M                                  & 90.88\%          \\
Majority Voting Ensemble~\cite{abouelnaga2017real}                                         & 120M                                 & 95.77\%          \\
GA Weighted Ensemble~\cite{abouelnaga2017real}                                             & 120M                                 & 95.98\%                                                                                                              \\ \hline
Original VGG-16~\cite{baheti2018detection}                                                  & 140M                                 & 94.44\%          \\
Regularized VGG-16~\cite{baheti2018detection}                                               & 140M                                 & 96.31\%          \\
Modified VGG-16~\cite{baheti2018detection}                                                  & 15M                                  & 95.54\%                                                                                                              \\ \hline
\begin{tabular}[c]{@{}l@{}}Pose-guided DenseNet~\cite{behera2018latent}\end{tabular} & 8.06M                                & 94.20\%          \\ \hline
\begin{tabular}[c]{@{}l@{}}MobileNet~\cite{howard2017mobilenets}\end{tabular}            & 4.20M                                & 94.67\%          \\ \hline
\begin{tabular}[c]{@{}l@{}}MobileNetV2~\cite{8578572}\end{tabular}          & 3.50M                                & 94.74\%          \\ \hline
\begin{tabular}[c]{@{}l@{}}NasNet Mobile~\cite{zoph2018learning}\end{tabular}        & 5.30M                                & 94.69\%          \\ \hline
\begin{tabular}[c]{@{}l@{}}SqueezeNet~\cite{iandola2016squeezenet}\end{tabular}           & 1.25M                                & 93.21\%          \\ \hline
\begin{tabular}[c]{@{}l@{}}MobileVGG~\cite{baheti2020towards}\end{tabular}            & 2.20M                                & 95.24\%          \\ \hline
VGG-one-attention~\cite{ai2019double}                                                & \textgreater{}140M                   & 84.82\%          \\
VGG-two-way-attention~\cite{ai2019double}                                             & \textgreater{}140M                   & 87.74\%                                                                                                           \\ \hline
\multicolumn{1}{l}{D-HCNN~\cite{qin2021distracted}}                                      & \multicolumn{1}{l}{0.76M}           & 95.59\%          \\ \hline
\multicolumn{1}{l}{\textbf{Teacher Network (Ours)}}             & \multicolumn{1}{l}{\textbf{44.62M}} & \textbf{96.35\%} \\ \hline
\multicolumn{1}{l}{\textbf{Student Network (Ours)}}             & \multicolumn{1}{l}{\textbf{0.42M}}  & \textbf{95.64\%} \\ \hline
\end{tabular}
\end{center}
\end{table}

\begin{figure}[t!]
\setlength{\fboxrule}{0.5pt}
\setlength{\fboxsep}{2pt}
\centering\includegraphics[width=\linewidth]{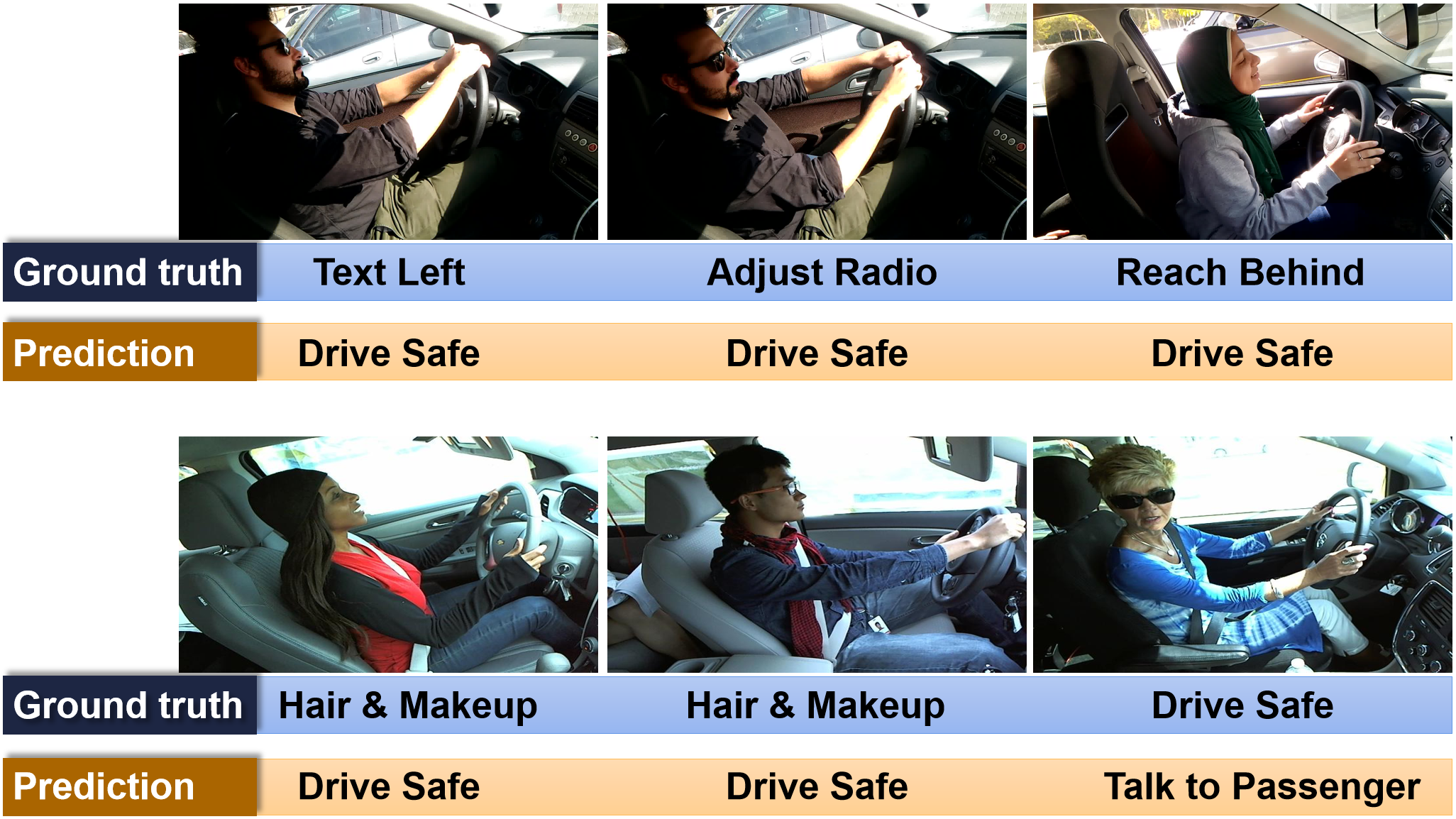}
\caption{Typical sample images that are wrongly classified by the networks proposed in this work. The ground-truth labels are confusing even for humans.}
\label{fig:wrong}
\end{figure}

\begin{table}[t!]
\centering
\caption{Comparison with state-of-the-art methods on SFD3} \label{tab:soa:sfd3}
\begin{threeparttable}
\begin{tabular}{lll}
\hline
\textbf{Approach}                                                                 & \textbf{\begin{tabular}[c]{@{}l@{}}Parameter\\ Size\end{tabular}} & \textbf{Accuracy} \\ \hline
\multicolumn{3}{c}{\cellcolor[HTML]{EFEFEF}Train:Test=7:3}        \\ \hline
VGG-16~\cite{dhakate2020distracted}                          & 140M                & 58.3\%    \\
VGG-19~\cite{dhakate2020distracted}                           & 142M                & 55.7\%    \\
Inception-V3~\cite{dhakate2020distracted}                     & 25.6M               & 92.90\%   \\
Inception-V3+Xception~\cite{dhakate2020distracted}            & 22.9M               & 82.50\%   \\
\begin{tabular}[c]{@{}l@{}}Inception-V3+Xception\\ +ResNet50+VGG-19~\cite{dhakate2020distracted} \end{tabular}  & 46.7M                                                             & 90.00\%                                                     \\ \hline
D-HCNN~\cite{qin2021distracted}                          & 0.76M               & 99.82\%   \\ \hline
\textbf{Teacher Network (Ours)} & \textbf{44.62M}     & \textbf{99.87$\pm$0.03\%} \\
\textbf{Student Network (Ours)} & \textbf{0.42M}      & \textbf{99.87$\pm$0.03\%} \\ \hline
\multicolumn{3}{c}{\cellcolor[HTML]{EFEFEF}Train:Test=7.5:2.5}    \\ \hline
Unpretrained VGG-16~\cite{masood2020detecting}             & 140M                & 99.43\%   \\
Pretrained VGG-16~\cite{masood2020detecting}               & 140M                & 99.57\%   \\
Unpretrained VGG-19~\cite{masood2020detecting}             & 142M                & 98.98\%   \\
Pretrained VGG-19~\cite{masood2020detecting}               & 142M                & 99.39\%                                                \\ \hline
MobileVGG~\cite{baheti2020towards}                       & 2.2M                & 99.75\%   \\ \hline
\begin{tabular}[c]{@{}l@{}}Transfer Learning\\ With ResNet~\cite{hssayeni2017distracted}\end{tabular}       & 60M                                                               & 85.00\%           \\ \hline
D-HCNN~\cite{qin2021distracted}                    & 0.76M               & 99.84\%   \\ \hline
\textbf{Teacher Network (Ours)} & \textbf{44.62M}     & \textbf{99.88$\pm$0.04\%} \\
\textbf{Student Network (Ours)} & \textbf{0.42M}      & \textbf{99.88$\pm$0.04\%} \\ \hline
\multicolumn{3}{c}{\cellcolor[HTML]{EFEFEF}Train:Test=8:2}        \\ \hline
Pixel SVC~\cite{zhang_2016}                       & -                   & 18.3\%    \\
SVC+HOG~\cite{zhang_2016}                         & -                   & 28.2\%    \\
SVC+PCA~\cite{zhang_2016}                         & -                   & 34.8\%    \\
SVC+Bbox+PCA~\cite{zhang_2016}                    & -                   & 40.7\%    \\
Original VGG-16~\cite{zhang_2016}                 & 140M                & 90.2\%    \\
VGG-GAP~\cite{zhang_2016}                        & 140M                & 91.3\%    \\
Original VGG-16+VGG-GAP~\cite{zhang_2016}         & 280M                & 92.6\%                                                \\ \hline
MLP~\cite{majdi2018drive}                             & -                   & 82.00\%   \\
RNN~\cite{majdi2018drive}                             & -                    & 91.7\%    \\
Drive-Net~\cite{majdi2018drive}                       & -                    & 95.00\%                                              \\ \hline
\begin{tabular}[c]{@{}l@{}}Vanilla CNN with Data\\ Transfer Learning~\cite{janet2020real}\end{tabular} & 26.05M                                                            & 97.05\%           \\
\begin{tabular}[c]{@{}l@{}}CNN with Data\\ Transfer Learning~\cite{janet2020real}\end{tabular}         & 3.5M                                                              & 71.72\%                                                     \\ \hline
HCF~\cite{huang2020hcf}                           & \textgreater{}72.3M & 96.74\%   \\ \hline
D-HCNN~\cite{qin2021distracted}                    & 0.76M               & 99.86\%   \\ \hline
\textbf{Teacher Network (Ours)} & \textbf{44.62M}     & \textbf{99.89$\pm$0.05\%} \\
\textbf{Student Network (Ours)} & \textbf{0.42M}      & \textbf{99.89$\pm$0.05\%} \\ \hline
\multicolumn{3}{c}{\cellcolor[HTML]{EFEFEF}Train:Test=9:1}        \\ \hline
AlexNet+SoftmaxLoss~\cite{okon2017detecting}             & 63.2M               & 96.80\%   \\
AlexNet+TripletLoss~\cite{okon2017detecting}             & 63.2M               & 98.70\%                                             \\ \hline
D-HCNN~\cite{qin2021distracted}               & 0.76M               & 99.87\%   \\ \hline
\textbf{Teacher Network (Ours)} & \textbf{44.62M}     & \textbf{99.91$\pm$0.05\%} \\
\textbf{Student Network (Ours)} & \textbf{0.42M}      & \textbf{99.91$\pm$0.05\%} \\ \hline
\end{tabular}
\begin{tablenotes}
\item[*] We illustrate the accuracy range of 10 random splits.
\end{tablenotes}
\end{threeparttable}
\end{table}

\section{Results and Discussions}

\subsection{Student Network Architecture Definition}

As mentioned above, we first train the super student network to approximate the prediction distribution of the teacher network. We carry out this experiment on the AUCD2, as it is a more challenging dataset than SFD3. The probability of choosing each candidate is shown in Table~\ref{tab:candidate_prob}. The candidates of the highest probability are marked in gray background. For convolutional layers, the searching guided by the teacher network chooses the third candidate for $b_1$ and the first candidate for all the other blocks. For pooling layers, the second candidate (max pooling) is selected for all the blocks. The reason might be that max pooling selects the brighter pixels or the features corresponding to the sharp pixels, and therefore more robust to illumination changes. 

Referring to Table~\ref{tab:candidates} and Table~\ref{tab:candidate_prob}, we define the architecture of the student network as Table~\ref{tab:candidate_results}. This architecture only requires 0.42M parameters. In the following experiments, we use this architecture as the student network on both datasets.

\subsection{Recognition Performance of the Teacher and Student Network}

In this subsection, we compare the recognition performance of the teacher network with and without progressive learning (PL), the student network trained from scratch and finetuned after transferring the teacher network's knowledge to the student network. The results are shown in Table~\ref{tab:performance}.

On the AUCD2, PL improves the teacher network by 1.06\%, which shows the effectiveness of PL. On the SFD3, the improvement brought by PL is small, which is 0.04\%--0.11\%. It is because backbone of the teacher has already achieved a high accuracy that is 99.75\%--99.87\%. Considering the very narrow possible improvement space, we suppose PL can be still regarded as effective on the SFD3. In the following experiments, we use the teacher network with PL to guide the search of the student network architecture and transfer knowledge to the student network. 

On both datasets, the student network trained from scratch already achieves a very high accuracy, which shows the architecture obtained by the proposed searching approach is effective for the DDR task. Knowledge distillation respectively improve 0.52\% and 0.03\%--0.05\% on the AUCD2 and SFD3, respectively. 

Considering that the accuracy for the datasets is almost saturated, it is interesting to see there is still room for the improvement by our proposed method.

In addition, since the AUCD2 dataset is somewhat unbalanced, we also show the F1-score obtained with this dataset in Table~\ref{tab:F_1_score}. PL improves the teacher network by 0.32\%--1.54\% in different categories. Knowledge distillation respectively improve 0.17\%--0.84\% for the student network in different categories.

\begin{table*}[t!]
\centering
\caption{The recognition accuracy of the 3D student network on the three splits of different tasks of the DAD~\cite{drive_and_act_2019_iccv}}\label{tab:drive_act_performance}
\setlength{\tabcolsep}{1.6mm}
\begin{tabular}{l|ll|ll|llllllll}
\hline
\multirow{3}{*}{} &
  \multicolumn{2}{c|}{\multirow{2}{*}{Fine-grained Activities}} &
  \multicolumn{2}{c|}{\multirow{2}{*}{Scenarios}} &
  \multicolumn{8}{c}{Atomic Action Units} \\ \cline{6-13} 
 &
  \multicolumn{2}{c|}{} &
  \multicolumn{2}{c|}{} &
  \multicolumn{2}{c|}{Action} &
  \multicolumn{2}{c|}{Object} &
  \multicolumn{2}{c|}{Location} &
  \multicolumn{2}{c}{Action-Object-Location} \\ \cline{2-13} 
 &
  Validation &
  Test &
  Validation &
  Test &
  Validation &
  \multicolumn{1}{l|}{Test} &
  Validation &
  \multicolumn{1}{l|}{Test} &
  Validation &
  \multicolumn{1}{l|}{Test} &
  Validation &
  Test \\
Split 1 &
  72.04\% &
  69.57\% &
  60.24\% &
  44.97\% &
  79.25\% &
  \multicolumn{1}{l|}{84.81\%} &
  71.59\% &
  \multicolumn{1}{l|}{65.50\%} &
  69.90\% &
  \multicolumn{1}{l|}{63.79\%} &
  44.89\% &
  51.22\% \\
Split 2 &
  67.75\% &
  66.08\% &
  46.04\% &
  47.68\% &
  81.16\% &
  \multicolumn{1}{l|}{79.77\%} &
  61.13\% &
  \multicolumn{1}{l|}{63.54\%} &
  66.48\% &
  \multicolumn{1}{l|}{61.27\%} &
  47.64\% &
  36.69\% \\
Split 3 &
  71.59\% &
  61.43\% &
  58.57\% &
  35.80\% &
  81.78\% &
  \multicolumn{1}{l|}{74.70\%} &
  66.22\% &
  \multicolumn{1}{l|}{46.96\%} &
  50.21\% &
  \multicolumn{1}{l|}{65.37\%} &
  41.15\% &
  41.08\% \\
Average &
  70.46\% &
  65.69\% &
  54.95\% &
  42.82\% &
  80.73\% &
  \multicolumn{1}{l|}{79.76\%} &
  66.31\% &
  \multicolumn{1}{l|}{58.67\%} &
  62.20\% &
  \multicolumn{1}{l|}{63.48\%} &
  44.56\% &
  43.00\% \\ \hline
\end{tabular}
\end{table*}

\begin{table*}[t!]
\caption{Comparison with state-of-the-art methods on the DAD~\cite{drive_and_act_2019_iccv}}\label{tab:drive_act_soa}
\setlength{\tabcolsep}{0.7mm}
\begin{tabular}{ll|l|llllllllllll}
\hline
\multicolumn{2}{l|}{\multirow{4}{*}{}} &
  \multirow{4}{*}{\begin{tabular}[c]{@{}l@{}}Parameter\\ Size\end{tabular}} &
  \multicolumn{12}{c}{Accuracy} \\ \cline{4-15} 
\multicolumn{2}{l|}{} &
   &
  \multicolumn{2}{c|}{\multirow{2}{*}{\begin{tabular}[c]{@{}c@{}}Fine-grained\\ Activities\end{tabular}}} &
  \multicolumn{2}{c|}{\multirow{2}{*}{Scenarios}} &
  \multicolumn{8}{c}{Atomic Action Units} \\ \cline{8-15} 
\multicolumn{2}{l|}{} &
   &
  \multicolumn{2}{c|}{} &
  \multicolumn{2}{c|}{} &
  \multicolumn{2}{c|}{Atomic Action} &
  \multicolumn{2}{c|}{Object} &
  \multicolumn{2}{c|}{Location} &
  \multicolumn{2}{c}{\begin{tabular}[c]{@{}c@{}}Action-Object\\ -Location\end{tabular}} \\ \cline{4-15} 
\multicolumn{2}{l|}{} &
   &
  Val &
  \multicolumn{1}{l|}{Test} &
  Val &
  \multicolumn{1}{l|}{Test} &
  Val &
  \multicolumn{1}{l|}{Test} &
  Val &
  \multicolumn{1}{l|}{Test} &
  Val &
  \multicolumn{1}{l|}{Test} &
  Val &
  Test \\ \cline{1-3}
\multicolumn{1}{l|}{\multirow{4}{*}{\begin{tabular}[c]{@{}l@{}}Body Pose \\ Representation\end{tabular}}} &
  Interior~\cite{drive_and_act_2019_iccv} &
  - &
  45.23\% &
  \multicolumn{1}{l|}{40.30\%} &
  35.76\% &
  \multicolumn{1}{l|}{29.75\%} &
  57.62\% &
  \multicolumn{1}{l|}{47.74\%} &
  51.45\% &
  \multicolumn{1}{l|}{41.72\%} &
  53.31\% &
  \multicolumn{1}{l|}{52.64\%} &
  9.18\% &
  7.07\% \\
\multicolumn{1}{l|}{} &
  Pose~\cite{drive_and_act_2019_iccv} &
  - &
  53.17\% &
  \multicolumn{1}{l|}{44.36\%} &
  37.18\% &
  \multicolumn{1}{l|}{32.96\%} &
  54.23\% &
  \multicolumn{1}{l|}{49.03\%} &
  49.90\% &
  \multicolumn{1}{l|}{40.73\%} &
  53.76\% &
  \multicolumn{1}{l|}{53.33\%} &
  8.76\% &
  6.85\% \\
\multicolumn{1}{l|}{} &
  Two-Stream~\cite{wang2017modeling} &
  - &
  53.76\% &
  \multicolumn{1}{l|}{45.39\%} &
  39.37\% &
  \multicolumn{1}{l|}{34.81\%} &
  57.86\% &
  \multicolumn{1}{l|}{48.83\%} &
  52.72\% &
  \multicolumn{1}{l|}{42.79\%} &
  53.99\% &
  \multicolumn{1}{l|}{54.73\%} &
  10.31\% &
  7.11\% \\
\multicolumn{1}{l|}{} &
  Three-Stream~\cite{martin2018body} &
  - &
  55.67\% &
  \multicolumn{1}{l|}{46.95\%} &
  41.70\% &
  \multicolumn{1}{l|}{35.45\%} &
  59.29\% &
  \multicolumn{1}{l|}{50.65\%} &
  55.59\% &
  \multicolumn{1}{l|}{45.25\%} &
  59.54\% &
  \multicolumn{1}{l|}{56.50\%} &
  11.57\% &
  8.09\% \\ \hline
\multicolumn{1}{l|}{\multirow{3}{*}{End-to-end}} &
  C3D~\cite{tran2015learning} &
  78.14M&
  49.54\% &
  \multicolumn{1}{l|}{43.41\%} &
  - &
  \multicolumn{1}{l|}{-} &
  - &
  \multicolumn{1}{l|}{-} &
  - &
  \multicolumn{1}{l|}{-} &
  - &
  \multicolumn{1}{l|}{-} &
  - &
  - \\
\multicolumn{1}{l|}{} &
  P3D ResNet~\cite{qiu2017learning} &
  65.74M &
  55.04\% &
  \multicolumn{1}{l|}{45.32\%} &
  - &
  \multicolumn{1}{l|}{-} &
  - &
  \multicolumn{1}{l|}{-} &
  - &
  \multicolumn{1}{l|}{-} &
  - &
  \multicolumn{1}{l|}{-} &
  - &
  - \\
\multicolumn{1}{l|}{} &
  I3D~\cite{carreira2017quo} &
  12.32M &
  69.57\% &
  \multicolumn{1}{l|}{63.64\%} &
  44.66\% &
  \multicolumn{1}{l|}{31.80\%} &
  62.81\% &
  \multicolumn{1}{l|}{56.07\%} &
  61.81\% &
  \multicolumn{1}{l|}{56.15\%} &
  47.70\% &
  \multicolumn{1}{l|}{51.12\%} &
  15.56\% &
  12.12\% \\ \hline
\multicolumn{2}{l|}{\textbf{3D Student Network (Ours)}} &
  \textbf{2.03M} &
  \textbf{70.46\%} &
  \multicolumn{1}{l|}{\textbf{65.69\%}} &
  \textbf{54.95\%} &
  \multicolumn{1}{l|}{\textbf{42.82\%}} &
  \textbf{80.73\%} &
  \multicolumn{1}{l|}{\textbf{79.76\%}} &
  \textbf{66.31\%} &
  \multicolumn{1}{l|}{\textbf{58.67\%}} &
  \textbf{62.20\%} &
  \multicolumn{1}{l|}{\textbf{63.48\%}} &
  \textbf{44.56\%} &
  \textbf{43.00\%} \\ \hline
\end{tabular}
\end{table*}

\subsection{Comparison with State-of-the-art Distracted Driver Recognition Approaches}

In this subsection, we compare our performance with the state-of-the-art approaches on AUCD2 and SFD3. 

Table~\ref{tab:soa:aucd2} shows the results on the AUCD3. The accuracy of the teacher network (96.35\%) surpasses the best previous accuracy (96.31\%), which is achieved by Regularized VGG-16~\cite{baheti2018detection}. Regularized VGG-16 has 140M parameters, whereas the teacher network in this work has 44.62M parameters (i.e., 31.87\% of the Regularized VGG-16 parameters), which shows the effectiveness of the teacher network on this dataset. The student network achieves 95.64\% with 0.42M parameters. For comparison, the original VGG-16 achieves 94.44\% with 140M parameters (i.e., 333.33 times of the student network parameters), and the modified VGG-16 achieves 96.54\% with 15M parameters (i.e., 35.71 times of the student network parameters)~\cite{baheti2018detection}.

Table~\ref{tab:soa:sfd3} shows the results on the SFD3. Both the teacher and student network achieve 99.86\%--99.91\%, which outperforms the best previous accuracy. The student network is recommended because it requires fewer parameters. 

D-HCNN~\cite{qin2021distracted} also achieves good accuracy on both datasets with small parameters. However, our student network is better because: (\romannumeral1) The student network has better accuracy than D-HCNN on both datasets; (\romannumeral2) The student network's parameters are only about 55.26\% of D-HCNN; (\romannumeral3) D-HCNN requires HOG images in addition to RGB images as input. Therefore, it needs to compute the HOG feature~\cite{dalal2005histograms} of every image when using D-HCNN, which is unfavorable for real-world applications.

Moreover, the student network has better real-time performance than other lightweight models. As shown in Table~\ref{tab:GFLOPS_TIME_COST}, for processing a single image in the test mode, the student network requires 2.25 GFLOPs takes 2.23 ms on 1080Ti + Intel i7-10700F. In comparison, MobileVGG~\cite{baheti2020towards} requires 2.11 GFLOPs and takes 5.19 ms. MobileNet~\cite{howard2017mobilenets} requires 0.59 GFLOPs and takes 3.54 ms. MobileNetV2~\cite{8578572} requires 0.33 GFLOPs and takes 6.94 ms. SqueezeNet~\cite{iandola2016squeezenet} requires 0.86 GFLOPs and takes 3.86 ms. D-HCNN~\cite{qin2021distracted} requires 31.10 GFLOPs and takes 7.40 ms. As D-HCNN requires HOG images as additional input, it takes additional 1.48ms per image to compute HOG for each image. Compared to previous lightweight networks, our network has no significant advantage in terms of GFLOPs but clearly has faster speed. It is because the parallelism of a convolutional network is mainly reflected in the calculation of each layer, and there is generally no parallelism across layers. So for convolutional neural networks used in high-speed DDR, large convolutional filter size is better than too deep layers. This fact was also pointed out by Qin~\textit{et al.}~\cite{qin2021distracted} and experimentally proved by them. Another advantage of our network is the aforementioned lower number of parameters, which allows our network to require less storage and memory space and be more easily deployed on in-vehicle devices.

Since GPUs can process multiple images in parallel, we also compare the time consumption of our network with other lightweight networks that process multiple images in parallel. For processing one batch of images (32 images) in the test mode, the student network takes 35.69 ms on 1080Ti + Intel i7-10700F. In comparison, MobileVGG~\cite{baheti2020towards} takes 122.34 ms. MobileNet~\cite{howard2017mobilenets} takes 53.15 ms. MobileNetV2~\cite{8578572} takes 44.72 ms. SqueezeNet~\cite{iandola2016squeezenet} takes 48.79 ms. D-HCNN~\cite{qin2021distracted} takes 40.67 ms. 

As the proposed teacher and student networks achieve very high accuracy on both image-based DDR datasets~\cite{statefarm,alotaibi2019distracted}, it is important to know what images cause the small number of recognition failures. Figure~\ref{fig:wrong} shows the typical failure cases of the wrongly-predicted images together with their ground-truth labels and the prediction given by the proposed networks (the teacher network or student network). Those failure cases are even confusing for humans.

\subsection{Extending the Student Network to 3D for the Video-based Distracted Driver Recognition}

The above experiments have proposed a lightweight yet powerful network architecture (i.e., the student network) for image-based DDR. In this subsection, we extend the student network into a spatial-temporal 3D network to evaluate whether on the video-based DDR dataset~\cite{drive_and_act_2019_iccv}, the 3D student network can retrace the success of the student network architecture proposed for the image-based DDR. This experiment is inspired by the experiments of Hara \textit{et al.}~\cite{hara2018can}, in which the researchers replaced the 2D layers (e.g., 2D convolutional layers, 2D batch normalization layers, etc.) of the ResNet architectures~\cite{he2016deep} with 3D layers (e.g., 3D convolutional layers, 3D batch normalization layers, etc.) and proved that using 3D ResNet architectures together with Kinetics~\cite{kay2017kinetics} can retrace the successful history of 2D CNNs on ImageNet~\cite{deng2009imagenet}. Following Hara \textit{et al.}~\cite{hara2018can}, we set the size of the third dimension of each 3D convolutional kernel to be the same as the size of the first and second dimensions. For example, a 2D convolutional kernel of a $3 \times 3$ kernel size is extended to a 3D convolutional kernel of a $3 \times 3 \times3$ kernel size.

We conducted comprehensive experiments to evaluate the performance of the 3D student network for all the tasks on the DAD. The specific accuracy of each split and the average accuracy over the three splits are shown in Table~\ref{tab:drive_act_performance}. The comparison results with the state-of-the-art approaches on the DAD are shown in Table~\ref{tab:drive_act_soa}. It can be observed that the 3D student network outperforms the state-of-the-art approaches by a significantly large margin in both validation and testing sets. Our approach is 0.89\%--29.00\% higher than the previous best accuracy in the validation set and 2.05\%--30.88\% higher than the previous best accuracy in the test set. Besides, the 3D student network has only 2.03M parameters and is much more lightweight than the state-of-the-art approaches. The parameter size of the 3D student network is only 16.48\% of the parameter size of C3D~\cite{tran2015learning}, 3.09\% of the parameter size of P3D ResNet~\cite{qiu2017learning}, 2.60\% of the parameter size of I3D~\cite{carreira2017quo}. Moreover, the student network has better real-time performance than those 3D convolutional neural networks. As shown in Table~\ref{tab:GFLOPS_TIME_COST}, for processing a single video clips, 3D student network requires 37.20 GFLOPs and takes 25.35 ms on 1080Ti + Intel i7-10700F. In comparison, C3D requires 38.55 GFLOPs and takes 25.57 ms. P3D ResNet requires 18.67 GFLOPs and takes 148.04 ms. I3D 27.90 GFLOPs and takes 61.86 ms. Similar to the case of 2D student network, 3D student network does not have clear advantage in terms of GFLOPs, but has a clearly faster speed.

For processing 8 video clips ($8\times16$ frames) in the test mode, the 3D student network takes 479.83 ms on 1080Ti + Intel i7-10700F. In comparison, C3D takes 1452.86 ms. P3D ResNet takes 983.22 ms. I3D takes 588.44 ms.

\subsection{Discussion on the Implication of the Proposed Framework on the ITS
applications}

The implication of our approach to applications is as follows:

\begin{itemize}
 
\item[-]We construct a powerful teacher network using progressive learning to increase robustness to illumination changes from shallow to deep layers of a backbone CNN. The classification accuracy of the teacher network exceeds that of all existing approaches and is well suited for the DDR applications that do not require a particularly small computational overhead but rather high accuracy.

\item[-]Using NAS and knowledge distillation, we generate an effective student network with the guidance of the teacher network. The student network can achieve high DDR accuracy and has less parametric count and inference time than any existing lightweight DDR networks. The student network is suitable for applications with high parametric and inference time requirements.

\item[-]We extend the student network into a spatial-temporal 3D network for performing DDR based on small video clips. The 3D student network has better DDR accuracy, smaller parameter size, and faster speed than the existing approaches. The 3D student network is suitable for applications developed based on video clips.

\item[-]Our proposed framework combining knowledge distillation and NAS has the potential to become a general DDR network design framework for different applications.
\end{itemize}

\section{Additional experiments}

\begin{table}[t!]
\caption{The recognition accuracy of the teacher and student network on the three additional datasets} \label{tab:add_performance}
\begin{center}
\setlength{\tabcolsep}{1.2mm}
\begin{tabular}{lllll}
\hline
        & \multicolumn{2}{l}{Teacher Network} & \multicolumn{2}{l}{Student Network} \\ \hline
 &
  Backbone &
  \begin{tabular}[c]{@{}l@{}}Backbone\\ +PL\end{tabular} &
  \begin{tabular}[c]{@{}l@{}}Train from\\  Scratch\end{tabular} &
  \begin{tabular}[c]{@{}l@{}}Finetune after\\ Distillation\end{tabular} \\ \hline
SLD2~\cite{DBLP:journals/corr/abs-2011-08927}     & 99.53\%           & 99.74\%          & 99.48\%           & 99.74\%          \\ 
Gesture2012~\cite{barczak2011new} & 100\%             & 100\%            & 100\%             & 100\%            \\ 
USED~\cite{4408872}   & 97.91\%           & 98.75\%          & 89.17\%           & 92.08\%          \\ \hline
\end{tabular}
\end{center}
\end{table}

\begin{table}[t!]
\caption{Comparison results on the three additional datasets} \label{tab:add_SOA}
\begin{center}
\begin{tabular}{lll}
\hline
\multicolumn{1}{l}{\textbf{Approach}} &
  \multicolumn{1}{l}{\textbf{\begin{tabular}[c]{@{}l@{}}Parameter\\ Size\end{tabular}}} &
  \textbf{Accuracy} \\ \hline
\multicolumn{3}{c}{\cellcolor[HTML]{EFEFEF}SLD2~\cite{DBLP:journals/corr/abs-2011-08927}}                                             \\ \hline
\begin{tabular}[c]{@{}l@{}}Contour SVM-based\\ digit-gesture recognition\end{tabular} &
  - &
  69.00\% \\
\begin{tabular}[c]{@{}l@{}}CNN-based digit-gesture\\ recognition\end{tabular} &
  1.84M &
  98.32\% \\
\multicolumn{3}{l}{\cite{saha2019bangla}}                             \\ \hline
Increasing Filter Size                                 & 5.46M           & 99.68\%          \\
Decreasing Filter Size                                 & 0.76M           & 99.68\%          \\
\multicolumn{3}{l}{\cite{qin2021distracted}}                                                                       \\ \hline
\textbf{Teacher Network (Ours)}                        & \textbf{44.62M} & \textbf{99.74\%} \\
\textbf{Student Network (Ours)}                        & \textbf{0.42M}  & \textbf{99.74\%} \\ \hline
\multicolumn{3}{c}{\cellcolor[HTML]{EFEFEF}Gesture2012~\cite{barczak2011new}}                                     \\ \hline
\begin{tabular}[c]{@{}l@{}}CNN-based digit-gesture\\ recognition\\ \cite{saha2019bangla}\end{tabular} &
  1.84M &
  100\% \\ \hline
Increasing Filter Size                                 & 5.46M           & 96.30\%          \\
Decreasing Filter Size                                 & 0.76M           & 94.10\%          \\
\multicolumn{3}{l}{\cite{qin2021distracted}}                                   \\ \hline
\textbf{Teacher Network (Ours)}                        & \textbf{44.62M} & \textbf{100\%}   \\
\textbf{Student Network (Ours)}                        & \textbf{0.42M}  & \textbf{100\%}   \\ \hline
\multicolumn{3}{c}{\cellcolor[HTML]{EFEFEF}USED~\cite{4408872}}                                            \\ \hline
\begin{tabular}[c]{@{}l@{}}SIFT+GGM~\cite{4408872}\end{tabular}   & -               & 73.4\%           \\ \hline
HMP                                                    & -               & 85.7\%           \\
SIFT+SC                                                & -               & 82.7\%           \\
\multicolumn{3}{l}{\cite{Bo2011Hierarchical}}                                                    \\ \hline
\begin{tabular}[c]{@{}l@{}}OB~\cite{li2014object}\end{tabular}         & -               & 76.3\%           \\ \hline
\begin{tabular}[c]{@{}l@{}}Places-CNN\end{tabular} & 61M             & 94.12\%          \\
ImageNet-CNN                                           & 61M             & 94.42\%          \\
Hybrid-CNN                                             & 122M            & 94.22\%          \\
\multicolumn{3}{l}{\cite{NIPS2014_3fe94a00}}                                                                       \\ \hline
\begin{tabular}[c]{@{}l@{}}TPN-FS~\cite{bai2017learning}\end{tabular}     & 121M            & 95.2\%           \\ \hline
DTCTH(LSVM)                                            & -               & 85.16\%          \\
DTCTH(HI)                                              & -               & 88.18\%          \\
\multicolumn{3}{l}{\cite{rahman2017dtcth}}                                              \\ \hline
CLGC(RGB-RGB)                                          & -               & 86.4\%           \\
CLGC(RGB-HSV)                                          & -               & 90.85\%          \\
\multicolumn{3}{l}{\cite{Kabbai2019image}}                                                                       \\ \hline
\textbf{Teacher Network (Ours)}                        & \textbf{44.62M} & \textbf{98.75\%} \\
\textbf{Student Network (Ours)}                        & \textbf{0.42M}  & \textbf{92.08\%} \\ \hline
\end{tabular}
\end{center}
\end{table}

We also evaluate our approach on three additional datasets, which are not for the DDR task but have the same characteristic: small diversity and strong inter-class similarity. The three additional datasets are Sign Language Digits Dataset (SLD2)~\cite{DBLP:journals/corr/abs-2011-08927}, Gesture Dataset 2012 (Gesture2012)~\cite{barczak2011new}, and UIUC Sports Event Dataset (USED)~\cite{4408872}. SLD2 and Gesture2012 are image datasets for hand sign language recognition, which are also used by Qin \textit{et al.}~\cite{qin2021distracted} as additional datasets to evaluate D-HCNN~\cite{qin2021distracted}. USED is an image dataset for sport event recognition.

On the three additional datasets, we compared the recognition performance of the teacher network with and without progressive learning (PL), the student network trained from scratch and finetuned after knowledge transferring. The results are shown in Table~\ref{tab:add_performance}. We also compared our approach with the state-of-the-art approaches on the three additional datasets, and the results are shown in the Table~\ref{tab:add_SOA}.

Both the teacher and student networks achieve 99.74\% on the SLD and 100\% on the Gesture2012, which reach state-of-the-art performance on the two datasets. The student network has much fewer parameters than other state-of-the-art approaches on these two datasets.

On the USED, the improvement brought by  PL and knowledge transfer is obvious. PL improves the teacher network by 0.84\% and knowledge transfer improves the student network by 2.91\%. The accuracy of the teacher network is 98.75\%, which surpasses the best previous accuracy by 3.55\%. The student network achieves 92.08\% with 0.42M parameters.

\section{Conclusion}
In this paper, we proposed a novel framework for distracted driver recognition to achieve high accuracy with a small number of parameters. This framework first builds a powerful teacher network based on progressive learning and then uses the teacher network to guide the searching of an optimal architecture for a student network, which is lightweight but can achieve high accuracy. Thereafter, the teacher network is used again to transfer the knowledge to the student network. The teacher network outperforms the previous state-of-the-art approaches on the Statefarm Distracted Driver Detection Dataset and AUC Distracted Driver Dataset. The student network achieves high accuracy with extremely tiny parameters on both datasets. The student network architecture can be extended into a spatial-temporal 3D convolutional neural network for recognizing distracted driving behaviors from video clips. The 3D student network significantly outperforms the previous state-of-the-art approaches with only 2.03M parameters on the Drive\&Act Dataset.


\bibliographystyle{IEEEtran}
\bibliography{egbib}

\end{document}